\documentclass[10pt,english,journal,compsoc]{IEEEtran}
\usepackage[latin9]{inputenc}
\usepackage{color}
\usepackage{babel}
\usepackage{float}
\usepackage{booktabs}
\usepackage{amsmath}
\usepackage{amssymb}
\usepackage{stmaryrd}
\usepackage{graphicx}
\usepackage[unicode=true,
 bookmarks=false,
 breaklinks=false,pdfborder={0 0 1},backref=section,colorlinks=false]
 {hyperref}

\makeatletter

\providecommand{\tabularnewline}{\\}
\floatstyle{ruled}
\newfloat{algorithm}{tbp}{loa}
\providecommand{\algorithmname}{Algorithm}
\floatname{algorithm}{\protect\algorithmname}

\let\oldforeign@language\foreign@language
\DeclareRobustCommand{\foreign@language}[1]{%
  \lowercase{\oldforeign@language{#1}}}

\usepackage{babel}
\usepackage{babel}
\usepackage{babel}
\usepackage{array}
\usepackage{multirow}
\usepackage{subfigure}
\usepackage{babel}
\usepackage{url}
\usepackage{algorithm}
\usepackage{algpseudocode}

\ifCLASSOPTIONcompsoc
  \usepackage[nocompress]{cite}
\else
  \usepackage{cite}
\fi

\makeatother

\begin{document}

\title{Learning to Augment Expressions for Few-shot Fine-grained Facial Expression Recognition}
\author{Wenxuan Wang, Yanwei Fu, Qiang Sun, Tao Chen, Chenjie Cao, Ziqi Zheng,
Guoqiang Xu, Han Qiu, Yu-Gang Jiang, Xiangyang Xue \IEEEcompsocitemizethanks{
\IEEEcompsocthanksitem Wenxuan Wang, Yu-Gang Jiang, Xiangyang Xue
are with the School of Computer Science, Shanghai Key Lab of Intelligent
Information Processing, Fudan University, Shanghai, China. Email:
\{wxwang19,ygj,xyxue\}@fudan.edu.cn. \IEEEcompsocthanksitem Yanwei~Fu
is with the School of Data Science, Shanghai Key Lab of Intelligent
Information Processing, Fudan University, Shanghai, China. Email:
yanweifu@fudan.edu.cn. Yanwei~Fu is the corresponding author. \IEEEcompsocthanksitem
Qiang Sun is with the Academy for Engineering \& Technology, Fudan
University, Shanghai, China. Email: 18110860051@fudan.edu.cn. \IEEEcompsocthanksitem
Tao Chen is with the School of Information Science and Technology,
Fudan University, Shanghai, China. Email: eetchen@fudan.edu.cn \IEEEcompsocthanksitem
Chenjie Cao, Ziqi Zheng, Guoqiang Xu, Han Qiu are with Ping An OneConnect,
Shanghai, China. Email: \{caochenjie948,zhengziqi356,xuguoqiang371,hannaqiu\}@pingan.com.}\thanks{}}

%


\markboth{Journal of \LaTeX\ Class Files,~Vol.~14, No.~8, August~2015}{Shell \MakeLowercase{\textit{et al.}}: Bare Demo of IEEEtran.cls for Computer Society Journals}

%



\IEEEtitleabstractindextext{ 
\begin{abstract}
Affective computing and cognitive theory are widely used in modern
human-computer interaction scenarios. Human faces, as the most prominent
and easily accessible features, have attracted great attention from
researchers. Since humans have rich emotions and developed musculature,
there exist a lot of subtle and fine-grained expressions in real-world
applications. However, it is extremely time-consuming to collect and
annotate a large number of facial images, of which some subtle expressions
may even require psychologists to correctly categorize them. To the
best of our knowledge, the existing expression datasets are only limited
to several basic facial expressions, which are not sufficient to support
our ambitions in developing successful human-computer interaction
systems. To this end, a novel Fine-grained Facial Expression Database
-- $\mathrm{F}^{2}$ED is contributed in this paper; and such a dataset
costs one year to be collected by us and annotated with the help of
psychological annotators. Totally, it includes more than 200k images
with 54 facial expressions from 119 persons. So far as we know, this
is the first large dataset to label faces with subtle emotion changes
for the recognition of facial expressions. Considering the phenomenon
of uneven data distribution and lack of samples is common in real-world
scenarios, we further evaluate several tasks of few-shot expression
learning by virtue of our $\mathrm{F}^{2}$ED, which are to recognize
the facial expressions given only few training instances. These tasks
mimic human performance to learn robust and general representation
from few examples. To address such few-shot tasks, we propose a unified
task-driven framework -- Compositional Generative Adversarial Network
(Comp-GAN) learning to synthesize facial images and thus augmenting
the instances of few-shot expression classes. Essentially, Comp-GAN
consists of two generators: one for editing faces with desired expression
and the other for changing the face posture; so it can generate many
realistic and high-quality facial images according to specified posture
and expression information while keeping the identity features. Extensive
experiments are conducted on $\mathrm{F}^{2}$ED and existing facial
expression datasets, \emph{i.e.,} JAFFE and FER2013, to validate the efficacy
of our $\mathrm{F}^{2}$ED in pre-training facial expression recognition
network and the effectiveness of our proposed approach Comp-GAN to
improve the performance of few-shot recognition tasks. 
\end{abstract}



}

\maketitle





\section{Introduction\label{sec:introduction}}

Affective Computing is one important research topic for
human-computer interaction \cite{rouast2019deep}. With the development of deep models
deployed on mobile devices, affective computing enables various applications
in psychology, medicine, security and education \cite{calvo2015oxford,gordon2016affective}.
In general, human eyes can easily recognize the facial expression;
but it is still a challenge for artificial intelligence
algorithms to effectively recognize the versatile facial emotional
expressions.

\begin{figure}
\centering{}\includegraphics[scale=0.22]{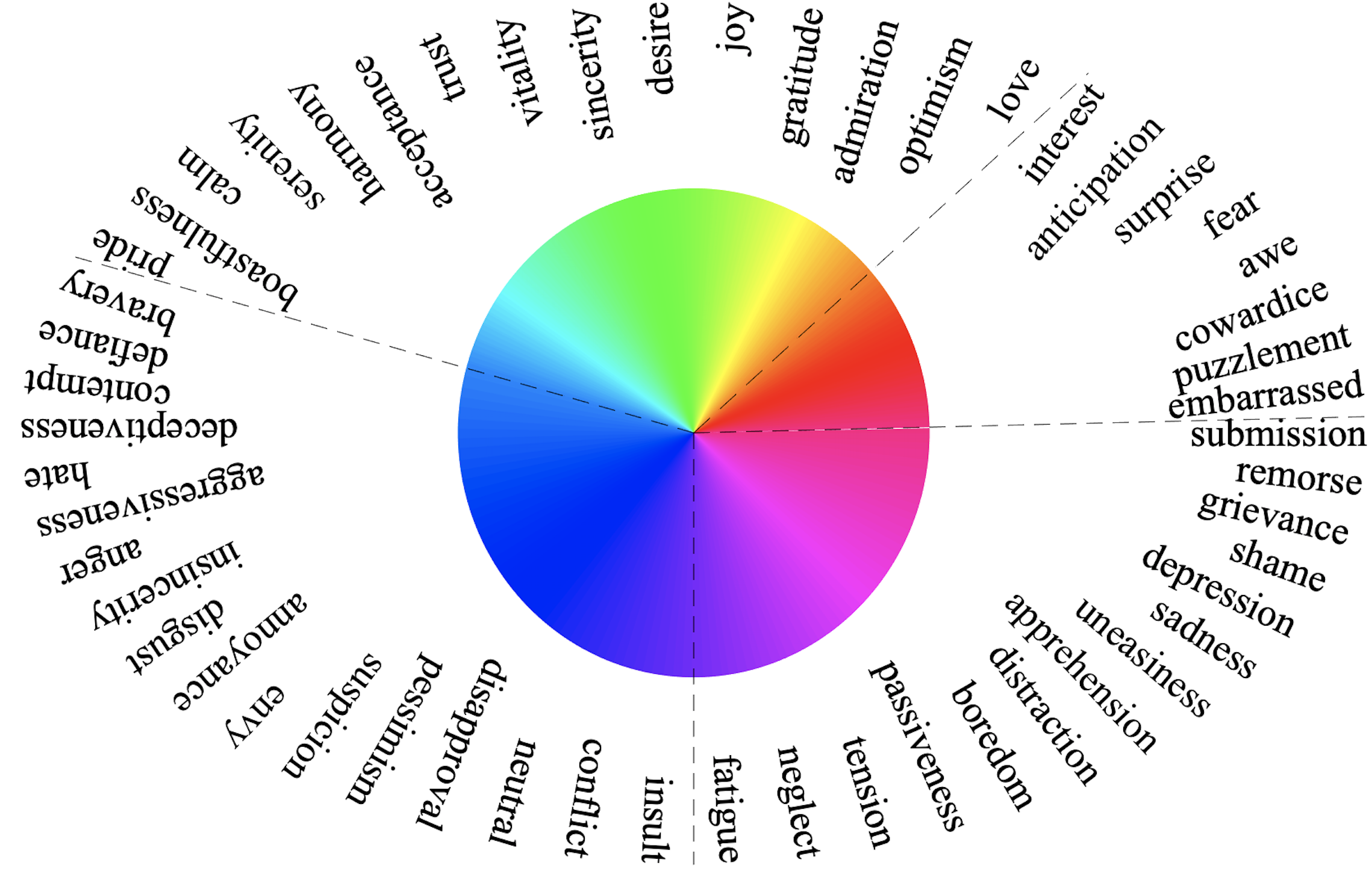} \caption{\label{fig:expressions} $\mathrm{F}^{2}$ED has 54 different facial
expression categories, which are organized into four large classes.}
\end{figure}

\begin{table*}
\centering{}\centering%
\begin{tabular}{c|c|c|c|c|c|c|c|c}
\hline 
{\small{}{}{}{}{}Dataset}  & {\small{}{}{}{}{}\#expression}  & {\small{}{}{}{}{}\#subject}  & {\small{}{}{}{}{}\#pose}  & {\small{}{}{}{}{}\#image}  & {\small{}{}{}{}{}\#sequence}  & {\small{}{}{}{}{}Resolution}  & {\small{}{}{}{}{}Pose list}  & {\small{}{}{}{}{}Condition}\tabularnewline
\hline 
\hline 
{\small{}{}{}{}{}CK+ \cite{kanade2000comprehensive}}  & {\small{}{}{}{}{}8}  & {\small{}{}{}{}{}123}  & {\small{}{}{}{}{}1}  & {\small{}{}{}{}{}327}  & {\small{}{}{}{}{}593}  & {\small{}{}{}{}{}$490\times640$}  & {\small{}{}{}{}{}F}  & {\small{}{}{}{}{}Controlled}\tabularnewline
\hline 
{\small{}{}{}{}{}JAFFE \cite{lyons1998coding}}  & {\small{}{}{}{}{}7}  & {\small{}{}{}{}{}10}  & {\small{}{}{}{}{}1}  & {\small{}{}{}{}{}213}  & {\small{}{}{}{}{}-}  & {\small{}{}{}{}{}$256\times256$}  & {\small{}{}{}{}{}F}  & {\small{}{}{}{}{}Controlled}\tabularnewline
\hline 
{\small{}{}{}{}{}KDEF \cite{lundqvist1998karolinska}}  & {\small{}{}{}{}{}7}  & {\small{}{}{}{}{}140}  & {\small{}{}{}{}{}5}  & {\small{}{}{}{}{}4,900}  & {\small{}{}{}{}{}-}  & {\small{}{}{}{}{}$562\times762$}  & {\small{}{}{}{}{}FL,HL,F,FR,HR}  & {\small{}{}{}{}{}Controlled}\tabularnewline
\hline 
{\small{}{}{}{}{}FER2013 \cite{fer2013}}  & {\small{}{}{}{}{}7}  & {\small{}{}{}{}{}-}  & {\small{}{}{}{}{}-}  & {\small{}{}{}{}{}35,887}  & {\small{}{}{}{}{}-}  & {\small{}{}{}{}{}$48\times48$}  & {\small{}{}{}{}{}-}  & {\small{}{}{}{}{}In-the-wild}\tabularnewline
\hline 
{\small{}{}{}{}{}FER-Wild \cite{mollahosseini2016facial}}  & {\small{}{}{}{}{}7}  & {\small{}{}{}{}{}-}  & {\small{}{}{}{}{}-}  & {\small{}{}{}{}{}24,000}  & {\small{}{}{}{}{}-}  & {\small{}{}{}{}{}-}  & {\small{}{}{}{}{}-}  & {\small{}{}{}{}{}In-the-wild}\tabularnewline
\hline 
{\small{}{}{}{}{}EmotionNet \cite{fabian2016emotionet}}  & {\small{}{}{}{}{}23}  & {\small{}{}{}{}{}-}  & {\small{}{}{}{}{}-}  & {\small{}{}{}{}{}100,000}  & {\small{}{}{}{}{}-}  & {\small{}{}{}{}{}-}  & {\small{}{}{}{}{}-}  & {\small{}{}{}{}{}In-the-wild}\tabularnewline
\hline 
{\small{}{}{}{}{}AffectNet \cite{mollahosseini2017affectnet}}  & {\small{}{}{}{}{}8}  & {\small{}{}{}{}{}-}  & {\small{}{}{}{}{}-}  & {\small{}{}{}{}{}450,000}  & {\small{}{}{}{}{}-}  & {\small{}{}{}{}{}-}  & {\small{}{}{}{}{}-}  & {\small{}{}{}{}{}In-the-wild}\tabularnewline
\hline 
\hline 
{\small{}{}{}{}{}$\mathrm{F}^{2}$ED}  & {\small{}{}{}{}{}54}  & {\small{}{}{}{}{}119}  & {\small{}{}{}{}{}4}  & {\small{}{}{}{}{}219,719}  & {\small{}{}{}{}{}5418}  & {\small{}{}{}{}{}$256\times256$}  & {\small{}{}{}{}{}HL,F,HR,BV}  & {\small{}{}{}{}{}Controlled}\tabularnewline
\hline 
\end{tabular}\caption{\label{tab:Comparison-with-existing}Comparison of $\mathrm{F}^{2}$ED
with existing facial expression database. In the pose list, F : front,
FL : full left, HL: half left, FR: full right, HR: half right, BV:
bird view.}
\end{table*}

It is well known that facial expression is the best visual representation
of a person's emotional status. According to \cite{ekman1994strong},
it is found that in years of observation and research the facial expression
of emotion is a common characteristic of human beings and contains
meaningful information in communication. Humans can always reliably
generate, understand and recognize facial emotional expressions. Indeed,
human emotional expressions are designed to deliver useful and reliable
information between different persons, so that people can decode each
other's psychological states from these designed emotion expressions.
Facial expression recognition is widely used in multiple applications
such as psychology, medicine, security, and education \cite{corneanu2016survey}.
In psychology, it can be used for depression recognition for analyzing
psychological distress. On the other hand, detecting the concentration
or frustration of students is also helpful in improving the educational approach.

Due to the above reasons, facial expression recognition has become the recent frontier
in affective computing and computer vision. Although facial expression
plays an important role in affective computing, there is no uniform
facial expression labeling system due to its subjective nature. According
to Ekman's theory \cite{ekman1999basic}, which is the most widely used labeling system in FER,
the emotion set is composed of six basic emotion types: anger, disgust,
fear, happy, sad and surprise. Plutchik's wheel \cite{plutchik1980emotion} expands the emotion
set to contain more diverse and \textit{subtle/fine-grained expressions},
which are very valuable to real-world applications. For example,
fatigue expression is important to monitor the status of drivers, 
which is critical for traffic safety.
Due to the simplicity of Ekman's theory, most academic datasets only
contain six basic emotions with an additional neutral emotion, such
as CK+ \cite{kanade2000comprehensive}, JAFFE \cite{lyons1998coding},
FER2013 \cite{fer2013} and FER-Wild \cite{mollahosseini2016facial},
as shown in Tab. \ref{tab:Comparison-with-existing}. Thus it is necessary
to create a dataset of more fine-grained emotions to fill the gap
between academic research and industrial applications.

\subsection{Fine-grained Expression Dataset}
Although, it is urgent to introduce fine-grained facial
expressions into the study, contributing such a large scale facial
expression dataset is non-trivial. Typically, the collection procedure should be
carefully designed to ensure that humans correctly convey the desired
facial expressions. Significant effort and contributions from both
psychologists and subjects have been made in our expression collection,
including explanations of emotion, scripts for emotion induction,
communication with psychologists, \emph{etc}. Furthermore, a careful
review mechanism \cite{kittur2008crowdsourcing} from the expert-level
judgments of psychologists is also designed to guarantee the quality
of collected facial expressions.

In this work, we contribute the first large fine-grained facial expression
dataset $\mathrm{F}^{2}$ED (Fine-grained Facial Expression Database)
with 54 expression emotions, such as calm, embarrassed, pride, tension
and so on, which includes abundant emotions with subtle changes, as
shown in Fig. \ref{fig:expressions}. These 54 expressions are classified
by referring to the recent psychological work \cite{Lee2017Reading}
with discernibility and rationality. Three psychologists and several doctoral
students participate in the whole collection and annotation process.
Further, we also consider the influence of facial pose changes on
the expression recognition, and introduce the pose as another attribute
for each expression. Four orientations (postures) including front,
half left, half right and bird view are labeled, and each has a balanced
number of examples to avoid distribution bias.

\begin{figure}
\centering{}\includegraphics[width=0.8\linewidth]{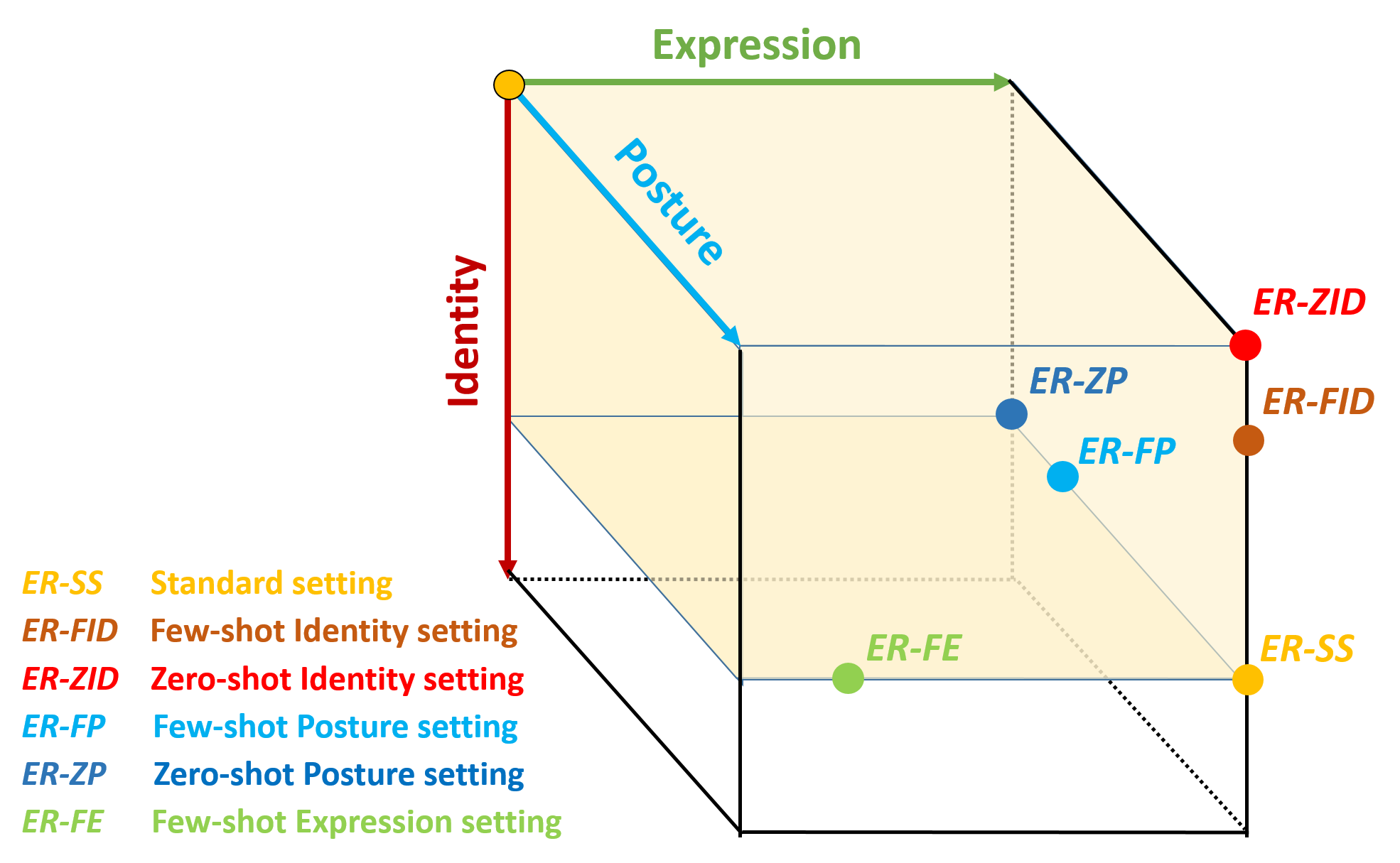}
\caption{The proposed several few-shot facial expression recognition learning tasks. \label{fig:setting}}
\end{figure}

\subsection{Few-shot Fine-grained Expression Recognition}

In the field of vision-based human-computer interaction, facial expression recognition (FER) is always 
a hot research topic. 
Recently, the renaissance of deep neural networks has significantly improved the
performance of FER tasks. The results on those well-known public facial
recognition datasets show that the deep neural networks based FER
methods which can learn both the low-level and high-level features
from facial images \cite{khorrami2015deep,mollahosseini2016going,minaee2019deep}
have outperformed the traditional methods based on hand-crafted
features \cite{kumar2009attribute,lowe2004distinctive,ojala2000gray,zhang2018joint}.

Despite the encouraging advancements in these FER works, several key challenges still remain in extending FER system to real-world
applications: (1) lack of sufficient and diverse high-quality training
data. (2) vulnerable to the variations of facial posture and person
identity.

Lacking sufficient data is a severe problem for FER, since deep neural
network needs a large scale labeled dataset to prevent the over-fitting
problem. Most research works frame the FER task as a typical supervised
learning problem, and assume there are plenty of training data for
each emotion. However, the annotation task for facial expression generally
requires devoted contributions from the experts, and the labeling
procedure is much more difficult and time-consuming than labeling
image class~\cite{deng2009imagenet}. It is thus a severe problem
in training deep FER models. To bypass the mentioned problem,
it is vital to get proper feature representations for classification
under the limited number of training samples. Typically, new expressions with only
few training examples may be encountered in the real world. Such few-shot
expression learning aims to mimic human performance in understanding
facial expressions from few training instances. For the first time,
this work extends the few-shot object classification to few-shot expression
learning, following the typical few-shot learning setting
\cite{lampert2014attribute}.
As illustrated in Fig. \ref{fig:setting}, we propose several novel learning tasks, and they are expression recognition
as follows,

\noindent (T1) Expression Recognition -- Standard Setting (ER-SS):  it is the standard supervised expression
classification task, which has relatively balanced training data for each identity with various poses and corresponding expressions.

\noindent (T2) Expression Recognition -- Few-shot IDentity setting (ER-FID): classifying expressions 
on the faces whose identity has only few training examples.

\noindent (T3) Expression Recognition --  Zero-shot IDentity setting (ER-ZID):  identifying the expressions from the faces whose identity is not in the training set.

\noindent (T4) Expression Recognition -- Few-shot Posture setting (ER-FP):  recognizing expressions by 
learning features from images of some specific poses which have been appeared only several times in the training data.

\noindent (T5) Expression Recognition --  Zero-shot Posture setting (ER-ZP): identifying expressions by utilizing the
features of faces that have specific poses not appeared in the training set.

\noindent (T6) Expression Recognition --  Few-shot Expression (ER-FE):  recognizing expressions by 
extracting discriminative features from few training data for novel expressions.

\subsection{Learning to Augment Faces}
The problem of few-shot learning is common in practical applications, and
the lack of training data can lead to a significant decrease in FER accuracy.
To alleviate the impediment of the expression-unrelated variations
such as poses, illuminations and identities, one approach is to use
various normalization mechanisms such as illumination normalization,
pose normalization \cite{qian2018pose,zhang2018joint}. However,
it is too cumbersome to develop different normalization mechanisms
for each expression-unrelated variation, and the mutual effects of
those normalization mechanisms may weaken the ability of deep FER
models. The data augmentation technique may mitigate this problem
in a more simple and smooth way. It can synthesize more diverse training
data to make the learned model more robust to the unrelated noise.
Several GAN-based methods are applied in synthesizing faces with different
expressions~\cite{yang2018identity}, poses~\cite{lai2018emotion}
and identities~\cite{chen2018vgan}, respectively. However, they
do not properly preserve the identity or expression information in
generating the target images, and the generated images are distorted.

In this work, we propose a novel unified Compositional Generative Adversarial
Network (Comp-GAN) to synthesize realistic facial images with arbitrary
poses and expressions while keeping the identity information. 
Our Comp-GAN model consists of two generators: one for generating images of desired expressions, and the other 
for editing the poses of faces. The two generators have different focuses and 
are complementary to each other. 
The structure of our Comp-GAN is composed of two branches, which have 
the same ultimate generating goal, thus forming a closed-loop to the network structure. 
Each branch has four generating steps, changing one attribute of the face at each step, 
and the goal of editing facial expression and posture 
is achieved through the successive multi-step generating process.
The difference between the two branches mainly lies in the different orders of 
generating attributes, \emph{e.g.}, one branch changes the posture first, 
while the other branch edits the expression first. 
The two branches constrain each other and improve the quality of the synthesized images.
We also apply a reconstruction learning process to re-generate the input image 
and encourage the generators for preserving the key information such as facial identity.
Aiming at enforcing the generative models to learn expression-excluding
details, we employ several task-driven loss functions to synthesize more
realistic and natural images, which can effectively solve the problem
of insufficient training data. With more diverse training images,
the FER model is more robust to various expression-unrelated changes. 

\vspace{0.1in}
\noindent \textbf{Contribution.}
The contributions are as follows, 
(1) For the first time, we introduce a new fine-grained facial expression 
dataset $\mathrm{F}^{2}$ED with three attributes
(face identity, pose, and expression) containing 54 different emotion types
and more than 200k examples. The 54 expressions have greatly enriched
the emotion categories, and provide more practical application scenarios.
(2) Considering the lack of diverse and sufficient training data 
in the real-word scenarios, we design
several few-shot expression learning tasks for FER to further investigate
how the poses, expressions, and subject identities affect the FER model performance.
(3) We propose a novel end-to-end Compositional Generative Adversarial
Network (Comp-GAN) to synthesize natural and realistic images to improve
the FER performance under the few-shot setting. 
We also introduce a closed-loop learning process
and several task-driven loss functions in Comp-GAN, which can encourage the
model to generate images with the desired expressions, specified poses
and meanwhile keep the expression-excluding details, such as identity
information. 
(4) We conduct extensive experiments on $\mathrm{F}^{2}$ED,
as well as JAFFE \cite{lyons1998coding} and FER2013 \cite{fer2013}
to evaluate our new dataset and framework. The experimental results
show that our dataset $\mathrm{F}^{2}$ED is large enough to be used
for pre-training a deep network to improve the recognition accuracy, and
the images generated by Comp-GAN can be used to alleviate the problem of insufficient
training data in few-shot expression setting, resulting in a more powerful model.

\section{Related Work}

\subsection{Affective Computing and Cognitive Theory}

Affective computing and cognitive theory are the intersection of psychology,
physiology, cognition, and computer technology \cite{calvo2015oxford}.
They can be widely applied to driver pressure or fatigue monitoring,
emotion robot \cite{nagama2018iot}, human-computer interaction (HCI)
\cite{rouast2019deep} and special medical service.

At present, extensive researches have been conducted on achieving affective
understanding and cognition between persons and computers \cite{picard2000affective}.
As the base signal of affective computing and cognitive theory, facial
expressions are the easiest visual features to be observed and detected.
Especially, the research of expression recognition is important for the research of HCI and emotional
robot \cite{zhiliang2006artificial}. Our work is primarily based
on the analysis and understanding of facial expressions to help affective
understanding and cognition.

\subsection{Evolutionary Psychology in Emotional Expression}

Ekman \textit{et al.} find \cite{ekman1994strong} that the expression 
of emotion is common to human beings in years of observation and research. No matter
where it is tested, humans can always reliably generate, understand
and recognize related emotional expressions. Indeed, human emotional
expressions are designed to provide information, and they need to
be delivered reliably, so that humans have coevolved automatic facial
expressions that decode these public expressions into insights of
other people's psychological states. Even though, people sometimes
lie, but inferences about emotional states from facial expressions
don't evolve unless they create a stronger advantage for the inferrer,
suggesting that these inferences are often valid and credible.

In recent years, psychologists and computational science specialists
have proposed expression recognition models based on cognition, probability
and deep learning network \cite{ortony1990cognitive,kshirsagar2002multilayer,khorrami2015deep}.
Most of these works are based on Ekman's six basic pan-cultural emotions
\cite{ekman1999basic}, however, human emotional world is rich and
colorful, with facial muscles and nerves well developed, so more and
more works begin to broaden the emotion and expression categories
that can be recognized \cite{lindquist2013hundred,xu2016heterogeneous}.
To further study the complex and subtle expressions of humans, inspired
by \cite{Lee2017Reading} which expands the emotion set, we collect
a new dataset $\mathrm{F}^{2}$ED with 54 subtle emotional expressions, thus
to a great extent to provide accurate and rich psychological and visual
intersection of expression information.

\subsection{Facial Expression Recognition in Computer Vision}

Original affective computing mainly focused on facial expression recognition
(FER), which has gained great progress in recent years. Facial expressions
can be recognized by two measurements: message and sign judgment~\cite{martinez2017automatic}.
In message judgment, facial expressions are categorized by the emotion
conveyed by the face such as angry, fear and happy. In signal judgment,
facial expressions are studied by physical signals such as raised
brows or depressed lips. There are mainly two kinds of FER methods 
according to the input type: static image FER
and dynamic sequence FER~\cite{li2018deep}. The static image FER
only uses the visual information in a single image to predict
the facial expression, whereas the dynamic sequence FER also leverages
the temporal information between the frames in videos to predict the
facial expression~\cite{li2018deep}. In this paper, we focus on
how to use a static image to predict the emotion type such as happy
and sad.


The most common static image-based FER method is composed of three
main steps: pre-processing, feature extraction and facial expression
classification. In the first step, there are two subtasks: face detection
and face alignment. For face detection, the faces are detected from
the image and labeled with bounding boxes. For face alignment, crucial landmarks
are used to align the face by warping affine method. In the second step,
feature extraction converts the image from pixel-level information
to high-level representation, such as appearance features (e.g. Garbor
wavelet~\cite{bartlett2005recognizing} LBP~\cite{shan2009facial},
HOG~\cite{dalal2005histograms} and SIFT~\cite{berretti20113d}),
geometric features and deep learning features. In the third step, an
additional classifier can be adopted in the facial expression classification,
such as MLP, SVM, and KNN.

Inspired by the success of deep neural networks on the vision tasks
such as image classification and object detection, extensive efforts\cite{khorrami2015deep,wang2017multi,zhang2016joint,mollahosseini2016going,minaee2019deep}
have been made to employ the deep neural networks in the FER tasks.
To name some promising works, Khorrami \emph{et al.} \cite{khorrami2015deep}
develop a zero-bias CNN for FER task, and find that those maximally
activated neurons in convolutional layers strongly correspond
to the Facial Action Units (FAUs)\cite{ekman1997face} by visualization.
In~\cite{mollahosseini2016going} a deep neural network with the inception
layer is proposed, and results show that the performance have achieved
or outperformed the state-of-the-art on MultiPIE, CK+, FER2013, and
other common datasets. Attentional CNN \cite{minaee2019deep} combines
a spatial transformer with CNN to focus on the most salient regions
of faces in FER.

\subsection{GAN-based Recognition Approach}

Generative Adversarial Networks (GANs) \cite{goodfellow2014generative}
based models have also been utilized in the FER task. Particularly,
GAN is a minimax game between a generator and a discriminator. 
Conditional Generative Adversarial Nets (cGAN)\cite{mirza2014conditional}
is proposed to generate the images conditioned on the class label.
Isola \emph{et al}. \cite{isola2017image} introduce a Pix2Pix model that
combines cGAN with U-Net to generate a new image conditioned on the
input image, low-level information is shared between input and output 
through U-Net. Zhu \emph{et al}.~\cite{zhu2017unpaired} propose a CycleGAN
model that employs a pair of GANs between two image domains to form
a cycle, cycle consistent loss is computed in both the forward cycle
and backward cycle. Qian \emph{et al.} \cite{qian2018pose} propose
a generative adversarial network (GAN) designed specifically for pose
normalization in re-id. Larsen \emph{et al}. advocate a VAE/GAN~\cite{larsen2015autoencoding}
that combines GAN with auto-encoder, and high-level features can be learned
and used for editing in the latent space of auto-encoder.

In order to weaken the impedance of expression-unrelated factors,
various GAN based FER methods are proposed. Yan \emph{et al.} \cite{yang2018facial}
propose a de-expression model to generate neutral expression images
from source images by cGAN\cite{mirza2014conditional},
then the residual information in intermediate layers is used for facial
expression. Lai \emph{et al}. in~\cite{lai2018emotion} propose a GAN to
generate a frontal face from a non-frontal face while preserving the
emotion. Yang \emph{et al}.~\cite{yang2018identity} utilize a cGAN
to produce six prototypic expression images for any source image,
and the expression of the source image is recognized by the minimum
distance between the source image and the generated six images in
a subspace. Chen \emph{et al}.~\cite{chen2018vgan} leverage a variational
generative adversarial network (VGAN) to encode the source image
into an identity-invariant latent space, and generate a new image with
desired identity code while keeping the expression unchanged. However,
these GAN based methods can only synthesize new images with one attribute different.

Several existing works~\cite{zhang2018joint,he2019attgan} attempt
to edit the multiple facial attributes in a unified model. He \emph{et al}.~\cite{he2019attgan}
propose an Attribute GAN(AttGAN) model which can edit any attribute
among a collection of attributes for face images by employing adversarial
loss, reconstruction loss and attribute classification constraints.
Zhang \emph{et al}. ~\cite{zhang2018joint} propose a joint pose and expression
GAN to generate new face images with different expressions under arbitrary
poses. However, these methods only employ a content similarity loss
on the cycle branch where the output shares the same attributes (\emph{e.g.,}
expression, pose) as the source image. Such design may be partially
due to the lack of target ground truth images in the training set.
Thus, the target branch that generates face images with different 
attributes is not constrained by the content similarity loss, which may 
weaken the ability to preserve the other facial information from the source image. 

\subsection{Few-shot Learning }

Few-shot learning \cite{fei2006one} aims to learn a new concept from
a limited number of labeled training data. There are three methods
commonly used in few-shot learning: meta-learning based methods\cite{finn2017model,ravi2016optimization},
metric-based methods \cite{snell2017prototypical} and augment-based methods\cite{zhang2018joint}. 
Meta-learning \cite{finn2017model} can transfer the knowledge from
previous different domains to boost the performance on the new task.
The pre-defined component in the training procedure can be taken as
prior knowledge, and trained by the meta-learner. For example, the
initial model parameters are taken as prior knowledge in MAML\cite{finn2017model},
and the parameter updating rules are taken as prior knowledge in Ravi's
work \cite{ravi2016optimization}. Inspired by FER tasks \cite{zhang2018joint},
our approach uses GAN to synthesize more training data rather than
linear transformation of pairwise images.

\begin{figure}
\centering{}\includegraphics[scale=0.36]{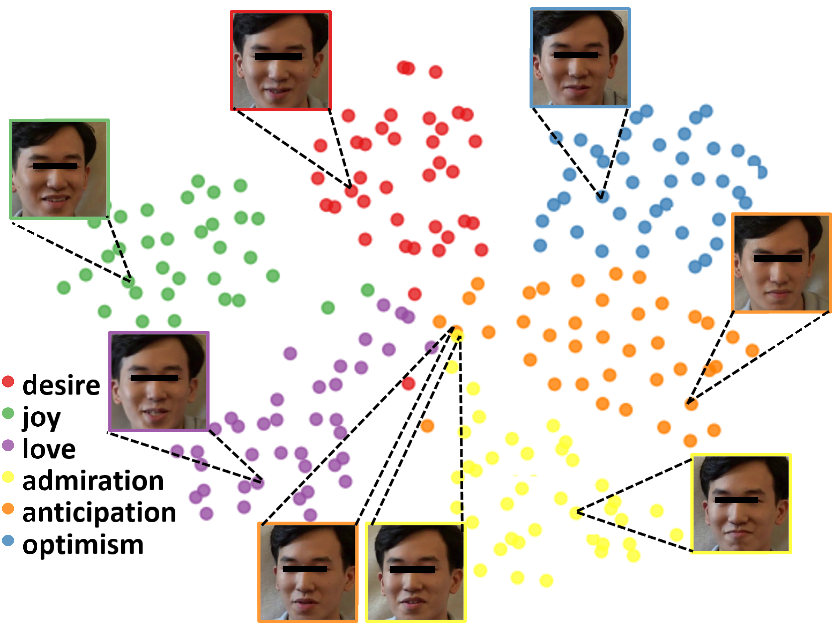}
\caption{\label{fig:t-sne} Visualization of 6 semantically indistinguishable
expressions of the same person using t-SNE.}
\end{figure}
 
\section{Fine-Grained Facial Expression Database}

\subsection{How to Differentiate Fine-grained Expressions?}

To further investigate the subtle expressions of the human faces,
we can classify expressions based on facial features (rather than
conceptual psychological states). We use this concept to construct
our dataset for two reasons.

First, the basis of expression in sensory functions means that certain
types of expressions are not arbitrary or random, and some expressions
look that way because they have interfaces that match their environment
\cite{darwin1998expression}. Thus, some indistinguishable mental
states \cite{susskind2008expressing} that are conceptually similar
(\emph{e.g}., fear is similar to disgust), present subtle expression
differences (\emph{e.g}., fear is opposite to disgust).

Second, we are studying subtle variations in facial expressions, which
have a wide range of real-world applications, physical attributes
(rather than conceptual attributes) are crucial because they constitute
essential signals to be sent to the recipient for understanding.

In this work, we expand the expression set to 54 types of expressions.
Particularly, in term of the theory of Lee \cite{Lee2017Reading},
which demonstrates the eye region can reliably convey diagnostic information
about discrete emotion states, \textit{e.g.}, the eye features associated
with happiness are consistent with a group of positive and stable
mental states, while the eye features associated with sadness align
with a cluster of negative and steady emotion states. To this end,
we can easily differentiate the expression set of 54 types of expressions,
which include more complex mental states based on seven eye features,
\textit{i.e.}, temporal wrinkles, wrinkles below eyes, nasal wrinkles,
brow slope, brow curve, brow distance, and eye apertures.

The 54 emotions can be clustered into 4 groups by the k-means clustering
algorithm as shown in Fig.~\ref{fig:expressions}, and the similar
mental-state map in \cite{Lee2017Reading} shows that the eye-narrowing
features of disgust are consistent with a range of mental states that
express social discrimination, such as hate, suspicion, aggression,
and contempt, which further prove the distinguishable nature of the
54 expressions.

We also visualize the feature distributions of data using randomly
sampled 6 kinds of indistinguishable mental expressions (\textit{i.e.,}
desire, joy, love, admiration, anticipation, and optimism) from the
same person via t-SNE in Fig. \ref{fig:t-sne}, which demonstrates
that our expressions are totally distinguishable. For the same person
with the same pose, the images with different expressions have higher
similarity, as the bottom two faces circled by orange and yellow dash
lines separately in Fig. \ref{fig:t-sne}, which also reflects the
difficulty of our dataset and the fine-grained expression recognition
task, and this is the main reason why we invite professional psychologists
to participate in the labeling work.

\subsection{Data Collection and Processing }

\noindent \textbf{Data Collection.} To make our dataset more practical, we invite three psychologists
and several doctoral students to conduct relevant research, determine
the categories of facial expressions, improve the process of guiding
participants and confirm the labeling methods. The whole video data
collection takes six months. Totally, we aim at capturing 54 different
types of expressions \cite{Lee2017Reading}, \emph{e.g.}, acceptance,
angry, bravery, calm, disgust, envy, fear, neutral and so on. We invite
more than 200 different candidates who are unfamiliar with our research
topics. Each candidate is captured by four cameras placed at four
different orientations to collect videos every moment. The four orientations
are front, half left, half right and bird view. The half left and
half right cameras have a horizontal angle of 45 degrees with the
front of the person, respectively. The bird view camera has a vertical
angle of 30 degrees with the front of the person. Each camera takes
25 frames per second.

The whole video capturing process is designed as a normal conversation
between the candidate and at least two psychological experts. The
conversation will follow a script which is calibrated by psychologists,
starting with the explanation of a particular expression definition
by psychologists, followed by a description of the relevant scene
including emotion, and finally letting the participants state similar
personal experiences to induce/motivate them to successfully express
the particular type of expression. For each candidate, we only select
5 seconds' video segment for each type of emotion, as noted and confirmed
by psychologists. To reduce the subjective interference of participants,
every subject has to cool down before a new emotion recording during
the data collection.

\vspace{0.1in}
\noindent  \textbf{Data Processing.} With gathered expression videos, we further
generate the final image dataset by human review, key image generation,
and face alignment. Specifically, the human review step is very important
to guarantee the general quality of recorded expressions. Three psychologists
and five doctoral students are invited to help us review the captured
emotion videos. Particularly, each captured video is given a score
of 1-3 by these experts based on the video quality. We only select
the videos that have an average score of beyond 1.5. Thus totally
about 119 identities' videos are kept finally. Then key frames are
extracted from each video. Face detection and alignment are conducted
by the toolboxes of Dlib and MTCNN \cite{zhang2016joint} over each
frame. Critically, the face bounding boxes are cropped from the original
images and resized to a resolution of $256\times256$ pixels. Finally,
we get the dataset $\mathrm{F}^{2}$ED of totally $219,719$ images
with 119 identities, 4 different views and 54 kinds of fine-grained
facial expressions.

\begin{figure}
\centering{}\includegraphics[scale=0.26]{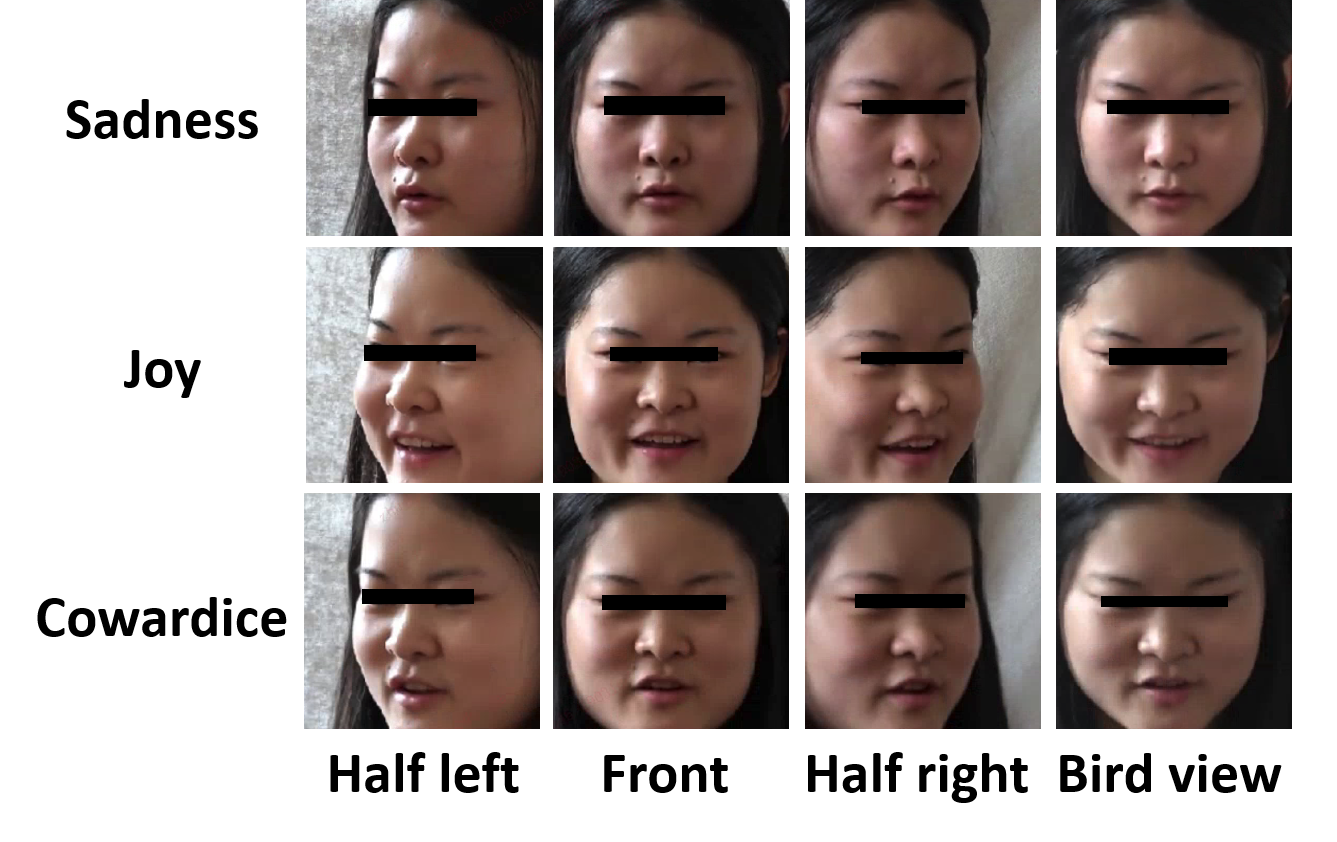}
\caption{\label{fig:dataset_samples} There are some facial examples of $\mathrm{F}^{2}$ED
with different poses and expressions.}
\end{figure}

\subsection{Statistics of $\mathrm{F}^{2}$ED }

Our dataset is labeled with identity, pose, and expression.

\vspace{0.1in}
\noindent \textbf{Identity.} The 119 persons are mainly university students
including 37 male and 82 female aging from 18 to 24. Each person expresses
emotions under the guidance and supervision of psychologists, and
the video is taken when the emotion is observed and confirmed by experts.

\begin{figure}
\centering{}\includegraphics[scale=0.7]{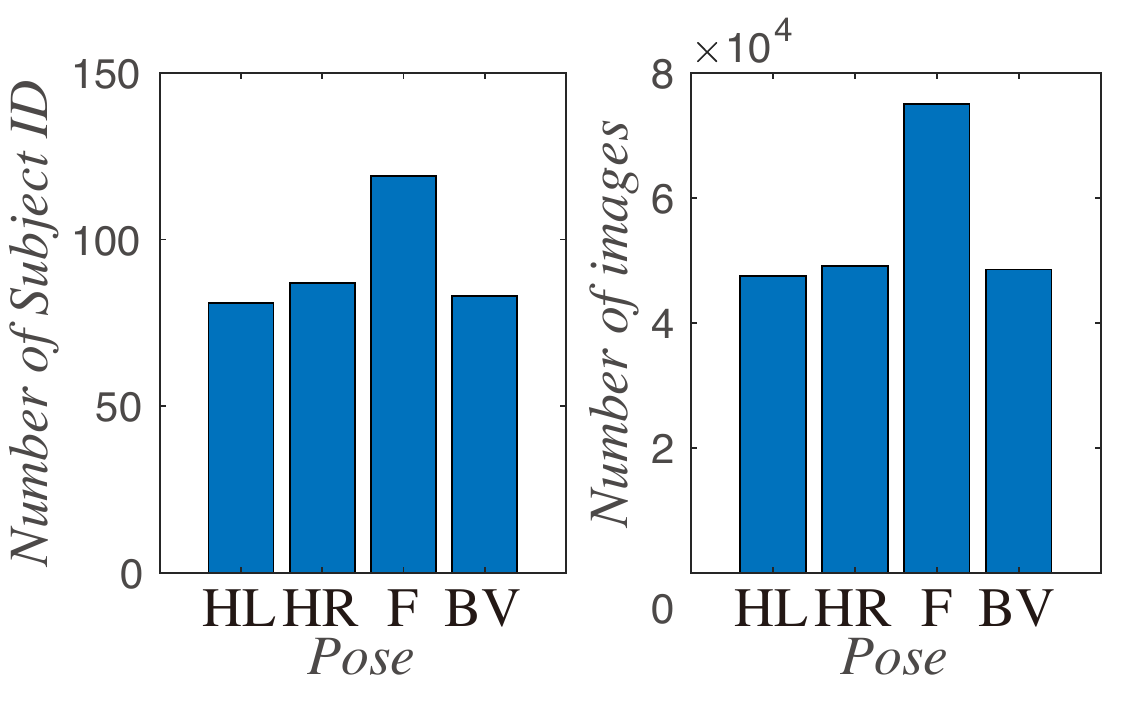} \caption{\label{fig:dist_pose} Data distribution on attribute of posture in $\mathrm{F}^{2}$ED.}
\end{figure}

\vspace{0.1in}
\noindent \textbf{Pose.} As an important type of meta-information, poses often
cause facial appearance changes. In real-world applications, facial
pose variations are mainly introduced by the relative position and
orientation changes of the cameras to persons. Fig.~\ref{fig:dataset_samples}
gives some examples of different poses. We collect videos from 4 orientations:
half left, front, half right and bird view, and we keep 47,053 half
left, 49,152 half right, 74,985 front, and 48,529 bird view images
in the $\mathrm{F}^{2}$ED, as shown in Fig. \ref{fig:dist_pose}.

\begin{figure*}
\centering{}\includegraphics[scale=0.43]{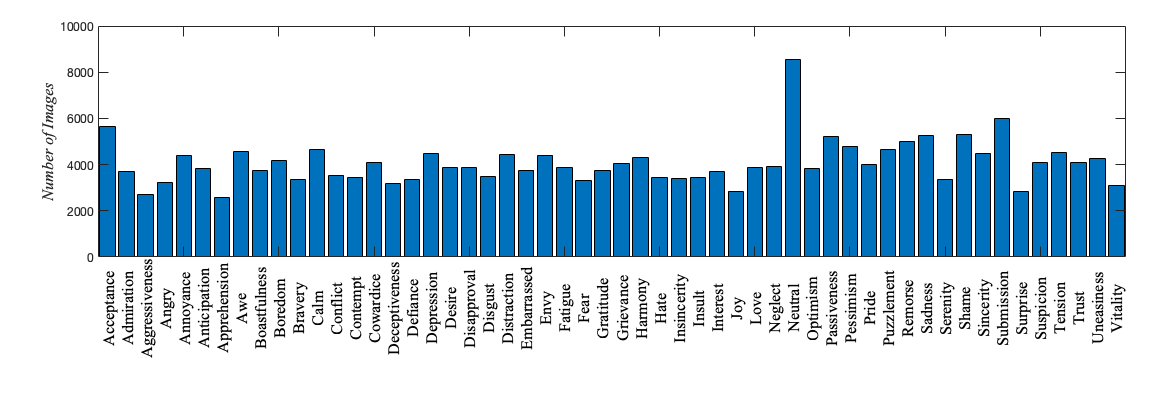}
\vspace{-0.15in}
\caption{\label{fig:dist_expression} Data distribution on attribute of expression in $\mathrm{F}^{2}$ED.}

\end{figure*}

\vspace{0.1in}
\noindent \textbf{Expression.} $\mathrm{F}^{2}$ED contains 54 fine-grained
facial expressions, which is helpful to understand the human emotion
status for affective computing and cognitive theory. The number of
images over each expression is shown in Fig. \ref{fig:dist_expression},
which indicates that $\mathrm{F}^{2}$ED has a relatively balanced
distribution across various expressions.

\vspace{0.1in}
\noindent \textbf{Comparison with Previous Datasets.} Table.~\ref{tab:Comparison-with-existing}
shows the comparison between our $\mathrm{F}^{2}$ED with existing
facial expression database. As shown in the table, our dataset contains
54 fine-grained expression types, while other datasets only contain
7 or 8 expression types in the controlled environment, 23 in the wild.
For the person number, CK+ \cite{kanade2000comprehensive}, KDEF \cite{lundqvist1998karolinska}
and $\mathrm{F}^{2}$ED are nearly the same. The current public facial
expression datasets are usually collected in two ways: in the wild
or in the controlled environment. The FER2013 \cite{fer2013}, FER-Wild
\cite{mollahosseini2016facial}, EmotionNet \cite{fabian2016emotionet},
AffectNet \cite{mollahosseini2017affectnet} are collected in the
wild, so the number of poses and subjects can not be determined. The
rest datasets are collected in a controlled environment, where the
number of poses for CK+ and JAFFE \cite{lyons1998coding} is 1, KDEF
is 5 and $\mathrm{F}^{2}$ED is 4. Our $\mathrm{F}^{2}$ED is the
only one that contains the bird view pose images which are very useful
in real-world applications. For image number, $\mathrm{F}^{2}$ED
contains 219,719 images, which is 44 times larger than the second-largest
dataset in the controlled environment. We show that our $\mathrm{F}^{2}$ED
is orders of magnitude larger than these existing datasets in terms
of expression class numbers, the number of total images and the diversity
of data.

\section{METHODOLOGY\label{sec:METHODOLOGY}}

\noindent \textbf{Problem Definition.} Typically, facial expression
recognition task (FER) aims to learn a classifier that can predict
the existence of expression to input images. Assume we have a training
dataset $\mathcal{D}_{s}=\left\{ \mathbf{\mathit{I}}_{j,(i,p,e)}\right\} _{j=1}^{N}$,
where $j$ means the $j$th image, $i$ is the identity label, $p$
indicates the posture label, and $e$ represents the expression label.
We use $\mathit{\mathbf{\mathit{I}}}_{(i,p,e)}$ to denote the face
image of the person $i$ with posture $p$ and expression $e$. Given
an unseen test face image $\mathbf{\mathit{I}^{\star}}_{(i,p,e)}$,
our goal is to learn a robust mapping function $\mathbf{\mathit{e}}^{\star}=\Psi\left(\mathbf{\mathit{I}^{\star}}_{(i,p,e)}\right)$
using all available training information to predict the expression
category $\mathbf{\mathit{e}}^{\star}$. To be noticed, each image
is only labeled with one expression type. 

\subsection{Few-shot Fine-grained Facial Expression Learning \label{subsec:Problem-Task}}
Generally due to the lack of sufficient
and diverse facial expression training data with dramatically changed
posture, learning a robust facial expression recognition model can
be very challenging. With regard to this problem, we introduce the
more practical few-shot learning case specializing to recognize the
samples appearing only a few or unseen during the training stage,
by learning the generalized features from a limited number of labeled data. 

Following the recent works~\cite{vinyals2016matching,Sachin2017,finn2017model,wang2018low},
we establish a group of few/zero-shot settings in FER task as in Fig.~\ref{fig:setting}: we firstly
define a base category set $C_{base}$ and a novel category set $C_{novel}$,
in which $C_{base}\cap C_{novel}=\phi$. Correspondingly, we have
a base dataset $D_{base}=\{(\mathbf{\mathit{I}}_{(i,p,e)}),(i,p,e)\subset C_{base}\}$,
and a novel dataset $D_{novel}=\left\{ (\mathbf{\mathit{I}}_{(i,p,e)}),(i,p,e)\subset C_{novel}\right\} $.
Our goal is to learn a generalized classification model that is able
to infer the novel class data trained on the $D_{base}$ and $D_{novel}$
with few or no samples per $C_{novel}$. Particularly, we propose the following 
tasks in the context of various standard problems, 

\noindent (T1) \emph{ER-SS} (FER under the Standard Setting): 
according to the general recognition task, the supervised setting is introduced 
into our work. We set $C_{novel}=\phi$ and $D_{novel}=\phi$, and 
directly learn the supervised classifier on the randomly sampled 80\% of all images, 
and test on the rest images. During the random sampling process, we ensure 
the expression, identity, and pose with a balanced distribution.

\noindent (T2) \emph{ER-FID} (FER under the Few-shot IDentity setting):
In the real-world applications, it is impossible to have training data 
with various expressions and postures from everyone.
So studying few-shot fine-grained expression recognition learning in terms of  
training identity is important to the real-world applications.
We randomly choose 20\% identities as $C_{novel}$ and the rest as
$C_{base}$, and randomly sample 1, 3 and 5 images per identity of $C_{novel}$
into $D_{novel}$ respectively;  80\%  of  total number of images
of $C_{base}$ as $D_{base}$. We stochastically choose 20\% of all images
of each identity from the rest images as the test data. In the above
random sampling process, the balance of expression and posture distribution
should be ensured simultaneously.

\noindent (T3) \emph{ER-ZID} (FER under the Zero-shot IDentity setting):
We randomly choose 20\% identities as $C_{novel}$ and the rest as
$C_{base}$, and sample no images per identity of $C_{novel}$
\textit{i.e.}, $D_{novel}=\phi$, 80\% of all images
of $C_{base}$ as $D_{base}$. We randomly select 20\% images of each
identity from the remaining data as the test samples. During the above
random splitting way, we also ensure the balance of expression and
pose distribution.

\begin{figure*}
\centering{}\includegraphics[width=0.7\linewidth]{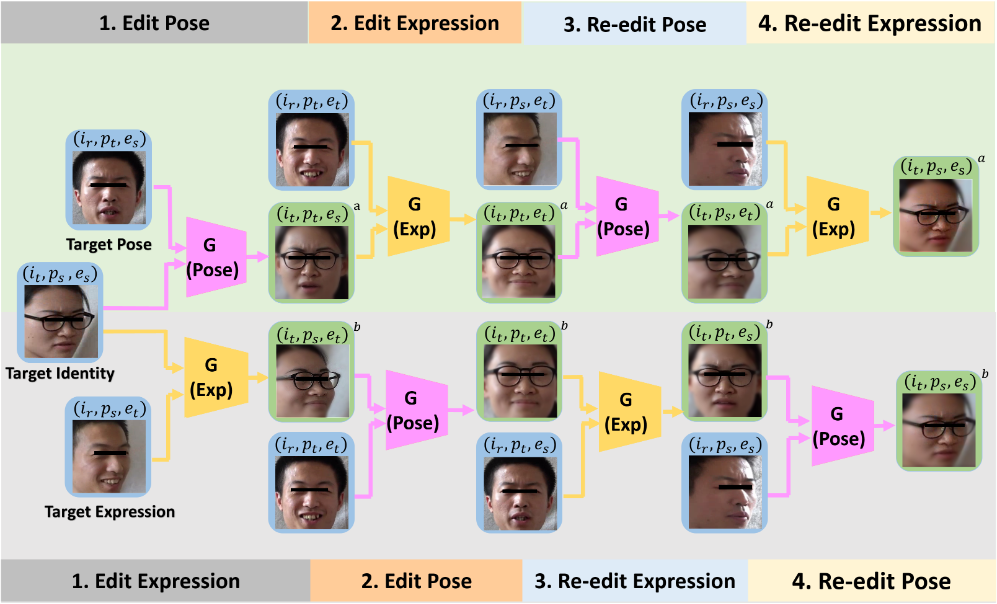}
\caption{The illustration of Comp-GAN framework. It is stacked by two components,
\emph{i.e.}, G (Exp) shown in yellow and G (Pose) as pink, and consists
of two branches with four steps. The images with blue borders represent
the input references, and the images with green borders are the generated
faces. The tags at the top of each image represent the identity,
posture and expression labels, (\textit{e.g.}, $(i_{t},p_{t},e_{s})^{a}$
means the generated image with target identity, target pose, and source expression
label from the upper branch). \label{fig:framework}}
\end{figure*}

\noindent (T4) \emph{ER-FP} (FER under the Few-shot Posture setting):
We choose left pose as $C_{novel}$ and the rest three poses as $C_{base}$,
\textit{i.e.}, right, front and bird-view, and randomly sample 1, 3, or
5 images with left pose into $D_{novel}$ individually; 80\% of all 
images of $C_{base}$ as $D_{base}$. We stochastically choose
20\% images of each facial pose category from the rest images as the
test data. In the above random sampling process, the balance of expression
and identity distribution should be ensured simultaneously.
Posture change has a significant impact on facial features, which greatly reduces 
the accuracy of expression recognition. Meanwhile, it is 
difficult to collect large training data with rich expressions and multiple 
poses. To overcome the lacking training samples in terms of poses, 
extract pose-invariant features is the key to solve the few-shot posture learning task.

\noindent (T5)\emph{ ER-ZP} (FER under the Zero-shot Posture setting):
We choose left pose as $C_{novel}$ and the rest three poses as $C_{base}$,
and sample no images with left posture as $D_{novel}$, \textit{i.e.},
 $D_{novel}=\phi$, 80\% of all images of
$C_{base}$ as $D_{base}$. We randomly select 20\% images of each
facial pose category from the remaining data as the test samples.
During the above random splitting way, we also ensure the expression
and identity into the balanced distribution.

\noindent (T6) \emph{ER-FE} (FER under the Few-shot Expression setting):
We randomly select 20\% expressions as $C_{novel}$ and the rest expressions
as $C_{base}$, and randomly sample 1, 3, or 5 images per expression of $C_{novel}$
into $D_{novel}$; 80\% of all images of $C_{base}$
as $D_{base}$. We stochastically choose 20\% images of each facial
pose category from the remaining data as the test samples. During
the above random splitting way, we simultaneously ensure the expression
and identity balanced distribution. Facial expressions vary a lot, so 
it is hard to collect a dataset with even distribution of expressions, 
which puts forward higher requirements for the application of expression
recognition in practice. Therefore, it is inevitable to study the 
fine-grained expression recognition task with uneven expression distribution.

\vspace{0.1in}
\noindent \textbf{Overview.} To tackle the challenges introduced in
Sec.~\ref{sec:introduction}, we propose the Compositional Generative
Adversarial Network (Comp-GAN) to generate desired expression and
specified pose images according to the input references while keeping
the expression-excluding details such as face identity. The generated
images can be adopted to train a robust expression recognition model.
The unified facial expression recognition architecture has
two components: Comp-GAN as in Fig.~\ref{fig:framework}, and expression
classifier network Comp-GAN-Cls based on LightCNN-29v2~\cite{Xiang2015A}.

\subsection{Structures of Comp-GAN\label{subsec:Comp-GAN framework}}

\noindent To solve the problem of changing postures and complex expressions
in the facial expression recognition task, we propose a Compositional
Generative Adversarial Network (Comp-GAN) to generate new realistic
face images, which dynamically edits the facial expression and pose
according to the reference image, while keeping the identity information.
As shown in Fig.~\ref{fig:framework}, a stacked GAN
with supervision information is presented to guide the generative
learning process. The generator in Comp-GAN is stacked by expression
and pose components, \emph{i.e.}, G (Exp) and G (Pose). The former
one serves as the editor of desired expressions, while the latter
one can synthesize faces by varying the facial posture.

Formally, we denote $i_{t}$, $i_{r}$, $p_{s}$, $p_{t}$, $e_{s}$
and $e_{t}$ as the target identity, reference identity, source posture,
target pose, source expression, and target expression, respectively.
The generator G (Pose) aims at transferring the source face posture
$p_{s}$ to the target pose $p_{t}$, and generator G (Exp) tries
to change the source facial expression $e_{s}$ to the target expression
$e_{t}$, while keeping their identity information $i_{t}$. Thus
our Comp-GAN can generate the target face $\mathbf{\mathit{I}}_{(i_{t},p_{t},e_{t})}$,
and an approximation to reconstruct the image $\mathbf{\mathit{I}}_{(i_{t},p_{s},e_{s})}$
to remain the pose-invariant and expression-invariant information.
Specifically, the whole model has two branches and four steps, and
the workflow of our Comp-GAN is illustrated in Alg. \ref{alg:Algorithm}.
Note that, we utilize the subindex $k\in\{a,b\}$ to indicate the intermediate
results produced by the $k$-th branch of Comp-GAN, and to better
understand, we simplify the $I_{(i_{t},p_{t},e_{s})}$ as $(i_{t},p_{t},e_{s})$
to represent the face with target identity, target pose and source
expression.

\begin{algorithm}
\textbf{Input:} $(i_{t},p_{s},e_{s})$ indicates the image with target
identity, $((i_{r},p_{t},e_{s}),(i_{r},p_{s},e_{t}),(i_{r},p_{t},e_{t}),(i_{r},p_{s},e_{s}))$
are the reference images with target posture or expression information.

\textbf{Output:} $((i_{t},p_{t},e_{s})^{a},(i_{t},p_{t},e_{t})^{b},(i_{t},p_{s},e_{t})^{a},(i_{t},p_{s},e_{s})^{b})$
are the generated images with edited posture by G (Pose); $((i_{t},p_{s},e_{t})^{b},(i_{t},p_{t},e_{t})^{a},(i_{t},p_{t},e_{s})^{b},(i_{t},p_{s},e_{s})^{a})$ mean the synthesized images with changed expression by G (Exp).

{\small{}1.}\vspace{-0.25in}
\begin{align*}
F_{P}:((i_{t},p_{s},e_{s}),(i_{r},p_{t},e_{s}))\rightarrow((i_{t},p_{t},e_{s})^{a})\text{,}\\
F_{E}:((i_{t},p_{s},e_{s}),(i_{r},p_{s},e_{t}))\rightarrow((i_{t},p_{s},e_{t})^{b})
\end{align*}

{\small{}2.}\vspace{-0.25in}
\begin{align*}
F_{E}:((i_{t},p_{t},e_{s})^{a},(i_{r},p_{t},e_{t}))\rightarrow((i_{t},p_{t},e_{t})^{a}),\\
F_{P}:((i_{t},p_{s},e_{t})^{b},(i_{r},p_{t},e_{t}))\rightarrow((i_{t},p_{t},e_{t})^{b})
\end{align*}

{\small{}3.}\vspace{-0.25in}
\begin{align*}
F_{P}:((i_{t},p_{t},e_{t})^{a},(i_{r},p_{s},e_{t}))\rightarrow((i_{t},p_{s},e_{t})^{a}),\\
F_{E}:((i_{t},p_{t},e_{t})^{b},(i_{r},p_{t},e_{s}))\rightarrow((i_{t},p_{t},e_{s})^{b})
\end{align*}

{\small{}4.}\vspace{-0.25in}
\begin{align*}
F_{E}:((i_{t},p_{s},e_{t})^{a},(i_{r},p_{s},e_{s}))\rightarrow((i_{t},p_{s},e_{s})^{a}),\\
F_{P}:((i_{t},p_{t},e_{s})^{b},(i_{r},p_{s},e_{s}))\rightarrow((i_{t},p_{s},e_{s})^{b})
\end{align*}

\caption{\label{alg:Algorithm}Algorithm of Comp-GAN.}
\end{algorithm}

In particular, $F_{P}$, \textit{i.e.}, G (Pose), learns to change
the pose while keeping the expression and identity information as
the reference data, and $F_{E}$, \textit{i.e.}, G (Exp), learns to
generate desired expression image while maintaining the expression-excluding
information. After the first two steps, we get the specified
pose data $(i_{t},p_{t},e_{s})^{a}$ and desired expression image
$(i_{t},p_{s},e_{t})^{b}$, as well as the $(i_{t},p_{t},e_{t})^{a}$
and $(i_{t},p_{t},e_{t})^{b}$ whose posture and expression are changed
simultaneously. We further utilize a reconstruction constraint to
re-generate the original faces using the same G (Pose) and G (Exp)
generators in the next two steps. It is worth noting that although
our generation model has four stages, it is only repeatedly built
with two generators.

\subsection{Loss Function in Comp-GAN \label{subsec:Comp-GAN loss}}

\noindent To synthesize realistic facial images, Comp-GAN consists of
the following losses: expression-prediction loss, ID-preserving loss,
posture-prediction loss, construction loss, reconstruction loss, closed-loop loss,
and adversarial loss. On $D_{base}$, we train a classifier
$F_{cls}$ based on LightCNN-29v2 \cite{Xiang2015A}, 
which can predict the expression, pose and identity label
simultaneously, to constrain the generative process. 
The classifier is further fine-tuned on the training
instances of the novel category set $C_{novel}$ for 
different few-shot learning tasks (T1) -- (T6).

\vspace{0.1in}
\noindent \textbf{Expression-prediction Loss.} We apply the classifier
to ensure the generated images with target expression, and employ
the cross-entropy loss $\mathit{\Phi}$ for model training to make
the learned features discriminative, 
\begin{center}
\begin{align}
L_{exp} & =\mathit{\Phi}(F_{cls}((i_{t},p_{t},e_{s})^{a}),e_{s})+\mathit{\Phi}(F_{cls}((i_{t},p_{t},e_{t})^{a}),e_{t})\nonumber \\
+ & \mathit{\Phi}(F_{cls}((i_{t},p_{s},e_{t})^{a}),e_{t})+\mathit{\Phi}(F_{cls}((i_{t},p_{s},e_{s})^{a}),e_{s})\nonumber \\
+ & \mathit{\Phi}(F_{cls}((i_{t},p_{s},e_{t})^{b}),e_{t})+\mathit{\Phi}(F_{cls}((i_{t},p_{t},e_{t})^{b}),e_{t})\nonumber \\
+ & \mathit{\Phi}(F_{cls}((i_{t},p_{t},e_{s})^{b}),e_{s})+\mathit{\Phi}(F_{cls}((i_{t},p_{s},e_{s})^{b}),e_{s})\label{eq: exp_loss}
\end{align}
\par\end{center}

\vspace{0.1in}
\noindent \textbf{Posture-prediction Loss.} The classifier is utilized
to constrain that the synthesized images have correct poses, with
cross-entropy loss $\mathit{\Phi}$ to train. The posture-prediction
loss is defined as, 
\begin{center}
\begin{align}
L_{pose} & =\mathit{\Phi}(F_{cls}((i_{t},p_{t},e_{s})^{a}),p_{t})+\mathit{\Phi}(F_{cls}((i_{t},p_{t},e_{t})^{a}),p_{t})\nonumber \\
+ & \mathit{\Phi}(F_{cls}((i_{t},p_{s},e_{t})^{a}),p_{s})+\mathit{\Phi}(F_{cls}((i_{t},p_{s},e_{s})^{a}),p_{s})\nonumber \\
+ & \mathit{\Phi}(F_{cls}((i_{t},p_{s},e_{t})^{b}),p_{s})+\mathit{\Phi}(F_{cls}((i_{t},p_{t},e_{t})^{b}),p_{t})\nonumber \\
+ & \mathit{\Phi}(F_{cls}((i_{t},p_{t},e_{s})^{b}),p_{t})+\mathit{\Phi}(F_{cls}((i_{t},p_{s},e_{s})^{b}),p_{s})\label{eq: pose_loss}
\end{align}
\par\end{center}

\vspace{0.1in}
\noindent \textbf{ID-preserving loss.} The cross-entropy loss $\mathit{\Phi}$
also is employed for identification to ensure the generated image
keeping the target identity information: 
\begin{center}
\begin{align}
L_{id} & =\mathit{\Phi}(F_{cls}((i_{t},p_{t},e_{s})^{a}),i_{t})+\mathit{\Phi}(F_{cls}((i_{t},p_{t},e_{t})^{a}),i_{t})\nonumber \\
+ & \mathit{\Phi}(F_{cls}((i_{t},p_{s},e_{t})^{a}),i_{t})+\mathit{\Phi}(F_{cls}((i_{t},p_{s},e_{s})^{a}),i_{t})\nonumber \\
+ & \mathit{\Phi}(F_{cls}((i_{t},p_{s},e_{t})^{b}),i_{t})+\mathit{\Phi}(F_{cls}((i_{t},p_{t},e_{t})^{b}),i_{t})\nonumber \\
+ & \mathit{\Phi}(F_{cls}((i_{t},p_{t},e_{s})^{b}),i_{t})+\mathit{\Phi}(F_{cls}((i_{t},p_{s},e_{s})^{b}),i_{t})\label{eq: id_loss}
\end{align}
\par\end{center}

\vspace{0.1in}
\noindent \textbf{Construction loss.} To generate realistic images,
we adopt the widely used strategy in generation task that using a
combination of $L1$ loss and perceptual loss \cite{johnson2016perceptual} on the pre-trained classifier
$F_{cls}$, which restricts the quality of produced image textures. 
\begin{center}
\begin{align}
L_{recon}^{1} & =\mid\mid(i_{t},p_{t},e_{s})^{a}-(i_{t},p_{t},e_{s})\mid\mid_{1}\nonumber \\
+ & \mid\mid(i_{t},p_{t},e_{t})^{a}-(i_{t},p_{t},e_{t})\mid\mid_{1}\nonumber \\
+ & \mid\mid(i_{t},p_{s},e_{t})^{b}-(i_{t},p_{s},e_{t})\mid\mid_{1}\nonumber \\
+ & \mid\mid(i_{t},p_{t},e_{t})^{b}-(i_{t},p_{t},e_{t})\mid\mid_{1}\label{eq: con_loss_1}
\end{align}
\par\end{center}

\begin{center}
\begin{align}
L_{con}^{2} & =\sum_{i=1}^{4}(\mid\mid F_{cls}^{i}((i_{t},p_{t},e_{s})^{a})-F_{cls}^{i}((i_{t},p_{t},e_{s}))\mid\mid_{1}\nonumber \\
+ & \mid\mid F_{cls}^{i}((i_{t},p_{t},e_{t})^{a})-F_{cls}^{i}((i_{t},p_{t},e_{s}))\mid\mid_{1}\nonumber \\
+ & \mid\mid F_{cls}^{i}((i_{t},p_{s},e_{t})^{b})-F_{cls}^{i}((i_{t},p_{s},e_{t}))\mid\mid_{1}\nonumber \\
+ & \mid\mid F_{cls}^{i}((i_{t},p_{t},e_{t})^{b})-F_{cls}^{i}((i_{t},p_{t},e_{t}))\mid\mid_{1})\label{eq: con_loss_2}
\end{align}
\par\end{center}

\noindent where $F_{cls}^{i}(I)$ indicates the feature map of image
$I$ of the $i$-th layer in $F_{cls}$. Finally, we define the reconstruction loss,
\begin{align}
L_{con}=\gamma_{1}L_{con}^{1}+\gamma_{2}L_{con}^{2}\label{eq:con_loss}
\end{align}

\noindent where $\gamma_{1}$and $\gamma_{2}$ are the trade-off parameters
for the $L1$ and the perceptual loss, respectively.

\vspace{0.1in}
\noindent \textbf{Reconstruction loss.} To capture more pose-invariant
and expression-invariant features for generating more natural images,
we introduce a reconstruction learning process to re-generate original
faces under the last two steps in Alg. \ref{alg:Algorithm}. We add reconstruction
loss as follows: 

\begin{align}
L_{recon}=\gamma_{1}L_{recon}^{1}+\gamma_{2}L_{recon}^{2}\label{eq:con_loss-1}
\end{align}

\noindent where $L_{recon}$ also contains $L1$ loss and perceptual
loss as:
\begin{center}
\begin{align}
L_{recon}^{1} & =\mid\mid(i_{t},p_{s},e_{t})^{a}-(i_{t},p_{s},e_{t})\mid\mid_{1}\nonumber \\
+ & \mid\mid(i_{t},p_{s},e_{s})^{a}-(i_{t},p_{s},e_{s})\mid\mid_{1}\nonumber \\
+ & \mid\mid(i_{t},p_{t},e_{s})^{b}-(i_{t},p_{t},e_{s})\mid\mid_{1}\nonumber \\
+ & \mid\mid(i_{t},p_{s},e_{s})^{b}-(i_{t},p_{s},e_{s})\mid\mid_{1}\label{eq: recon_loss_1}
\end{align}
\par\end{center}

\begin{center}
\begin{align}
L_{recon}^{2} & =\sum_{i=1}^{4}(\mid\mid F_{cls}^{i}((i_{t},p_{s},e_{t})^{a})-F_{cls}^{i}((i_{t},p_{s},e_{t}))\mid\mid_{1}\nonumber \\
+ & \mid\mid F_{cls}^{i}((i_{t},p_{s},e_{s})^{a})-F_{cls}^{i}((i_{t},p_{s},e_{s}))\mid\mid_{1}\nonumber \\
+ & \mid\mid F_{cls}^{i}((i_{t},p_{t},e_{s})^{b})-F_{cls}^{i}((i_{t},p_{t},e_{s}))\mid\mid_{1}\nonumber \\
+ & \mid\mid F_{cls}^{i}((i_{t},p_{s},e_{s})^{b})-F_{cls}^{i}((i_{t},p_{s},e_{s}))\mid\mid_{1})\label{eq: recon_loss_2}
\end{align}
\par\end{center}

\vspace{0.1in}
\noindent \textbf{Closed-Loop loss.} The two branches in Comp-GAN
are formed as a closed-loop, to balance the learning process and constrain
the generation between the branches. The closed-loop loss is proposed
to ensure the properties of faces with the same identity, pose and expression
between the two branches as similar as possible, and improve the ability of the model to find
potential identical features. So we define the closed-loop loss as, 
\begin{center}
\begin{align}
L_{loop} & =\mid\mid(i_{t},p_{t},e_{s})^{a}-(i_{t},p_{t},e_{s})^{b}\mid\mid_{1}\nonumber \\
+ & \mid\mid(i_{t},p_{s},e_{t})^{a}-(i_{t},p_{s},e_{t})^{b}\mid\mid_{1}\label{eq: loop_loss}
\end{align}
\par\end{center}

\vspace{0.1in}
\noindent \textbf{Comp-GAN loss.} Our two generators -- G (Pose)
and G (Exp) are followed by a discriminator $D$ that tries to detect
the synthesized faces to help improve the quality of the generated
image. The adversarial learning between the generator and discriminator
is introduced to make the generated images visually realistic. So we
define the adversarial loss $L_{adv}$ as: 
\begin{align}
\underset{G}{\mathrm{min}}\underset{D}{\mathrm{max}}\,\mathcal{L}_{GAN} & =\mathbb{E}_{\mathbf{\mathit{I}_{\mathit{input}}}\sim p_{d}\left(\mathbf{\mathit{I}}_{input}\right)}\left[\mathrm{log}\,D\left(\mathbf{\mathit{I}_{\mathit{input}}}\right)\right]\nonumber \\
+ & \left[\mathrm{log}\,\left(1-D\left(G_{Comp-GAN}\left(\mathbf{\mathbf{\mathit{I}_{\mathit{input}}},\mathbf{\mathit{I}_{\mathit{target}}}}\right)\right)\right)\right]\label{eq: adv_loss}
\end{align}

\noindent where $\mathbf{\mathbf{\mathit{I}_{\mathit{input}}}}$ means
the input image, and $\mathbf{\mathbf{\mathit{I}_{\mathit{target}}}}$
is the corresponding ground-truth image.

Generator G (Pose) targets at changing the facial posture and keeps
the expression and identity information, and Generator G (Exp) is
aiming to edit the expression of faces while maintaining the pose
and identity details, so the G (Pose) loss function and the G (Exp)
loss function are defined as: 
\begin{center}
\begin{align}
L_{G(Pose)} & =\lambda_{exp}L_{exp}+\lambda_{pose}L_{pose}+\lambda_{id}L_{id}+\lambda_{con}L_{con}\nonumber \\
 & +\lambda_{recon}L_{recon}+\lambda_{adv}L_{adv}+\lambda_{loop}L_{loop}\label{eq:Comp-GAN_Pose}
\end{align}
\par\end{center}

\begin{center}
\begin{align}
L_{G(Exp)} & =\mu_{exp}L_{exp}+\mu_{pose}L_{pose}+\mu_{id}L_{id}+\mu_{con}L_{con}\nonumber \\
 & +\mu_{recon}L_{recon}+\mu_{adv}L_{adv}+\mu_{loop}L_{loop}\label{eq:Comp-GAN_Exp}
\end{align}
\par\end{center}

\noindent where $\lambda_{exp}$, $\lambda_{pose}$, $\lambda_{id}$,
$\lambda_{con}$, $\lambda_{recon}$, $\lambda_{adv}$, and $\lambda_{loop}$
are the weights for the corresponding terms of $L_{G(Pose)}$, respectively;
and $\mu_{exp}$, $\mu_{pose}$, $\mu_{id}$, $\mu_{con}$, $\mu_{recon}$,
$\mu_{adv}$ and $\mu_{loop}$ are the weights for the corresponding
terms of $L_{G(Exp)}$, individually.

The loss functions we proposed above are critical to our generation
model. The expression-prediction loss, ID-preserving loss, and posture-prediction
loss are intuitive, which are used to constrain the correct attribute labels
of the synthesized faces and motivate the generation process to accurately
edit the pose and expression while maintaining its identity information.
The proposed construction loss and reconstruction loss both consist
of $L1$ loss and perceptual loss to constrain the generated images
to be as similar to the ground-truth. The special closed-loop loss
is based on our special network structure, which has two branched
and there are intersections among the generation aims, to encourage
the generative model to capture more potential identical features. The
last adversarial loss is a commonly used method in adversarial generative
networks to play the minimax game.

\section{Experiments}

\subsection{Results of Supervised Expression Recognition}

\subsubsection{Dataset and Setting}

\noindent Extensive experiments are conducted on JAFFE \cite{lyons1998coding} 
and FER2013 \cite{fer2013} to evaluate our proposed dataset $\mathrm{F}^{2}$ED
under the supervised expression learning task.

\noindent \textbf{Dataset}.\textbf{ }\textit{(1) JAFFE.} The Japanese
Female Facial Expression (JAFFE) database \cite{lyons1998coding}
contains 213 images of 256$\times$256 pixels resolution. The images
are taken from 10 Japanese female models in a controlled environment.
Each image is rated by 60 Japanese subjects with one of the following
seven emotion adjectives: natural, angry, disgust, fear, happy, sad
and surprise.\textit{ (2) FER2013}. The Facial Expression Recognition
2013 database \cite{fer2013} contains 35,887 gray-scale images of
48$\times$48 resolution. Most images are taken in the wild setting
which means more challenging conditions such as occlusion and pose
variations are included. They are labeled as one of the seven emotions
as described above. The dataset is split into 28,709 training images,
3,589 validation images and 3,589 test images.

\noindent \textbf{Settings}. Following the setting of \cite{minaee2019deep},
we conduct the experiments on FER2013 by using the entire 28,709 training
images and 3,589 validation images to train and validate our model,
which is further tested on the rest 3,589 test images. As for JAFFE,
we follow the split setting of the deep-emotion paper \cite{minaee2019deep}
to use 120 images for training, 23 images for validation, and keep
70 images for the test (7 emotions per face ID). In our $\mathrm{F}^{2}$ED,
we randomly choose 175,000 and 44,719 images for train and test, respectively.
The FER classification accuracy is reported as the evaluation metric
to compare different competitors. The results are averaged and reported
over multiple rounds.

\noindent \textbf{Competitors}. On FER2013, we compare against several
competitors, including Bag of Words~\cite{ionescu2013local}, VGG+SVM~\cite{georgescu2018local},
Go-Deep\emph{~}\cite{mollahosseini2016going}, DNNRL~\cite{guo2016deep},
Attention CNN~\cite{minaee2019deep}, and BOVW + local SVM \cite{georgescu2019local}.
These investigated classifiers are based on hand-crafted features,
or specially designed for FER. As for JAFFE, we compare with several
methods that are tailored for the tasks of FER, including Fisherface
\cite{abidin2012neural}, Salient Facial Patch \cite{happy2015automatic},
CNN+SVM \cite{shima2018image} and Attention CNN~\cite{minaee2019deep}.

\noindent \textbf{Implementation Details.} We train the baseline classification
network Comp-GAN-Cls based on LightCNN-29v2~\cite{Xiang2015A} using the SGD optimizer
with a momentum of 0.9 and decreasing the learning rate by 0.457 after
every 10 steps. The maximum epoch number is set to 50. The learning
rate and batch size vary depending on the dataset size, so we set
the learning rate/batch size as 0.01/128, 2$e-3$/64 and $5e-4$/32,
on $\mathrm{F}^{2}$ED, FER2013, and JAFFE, respectively.

\noindent 
\begin{figure}
\centering{}%
\begin{tabular}{cc}
\hspace{-0.3in}\includegraphics[width=0.23\textwidth]{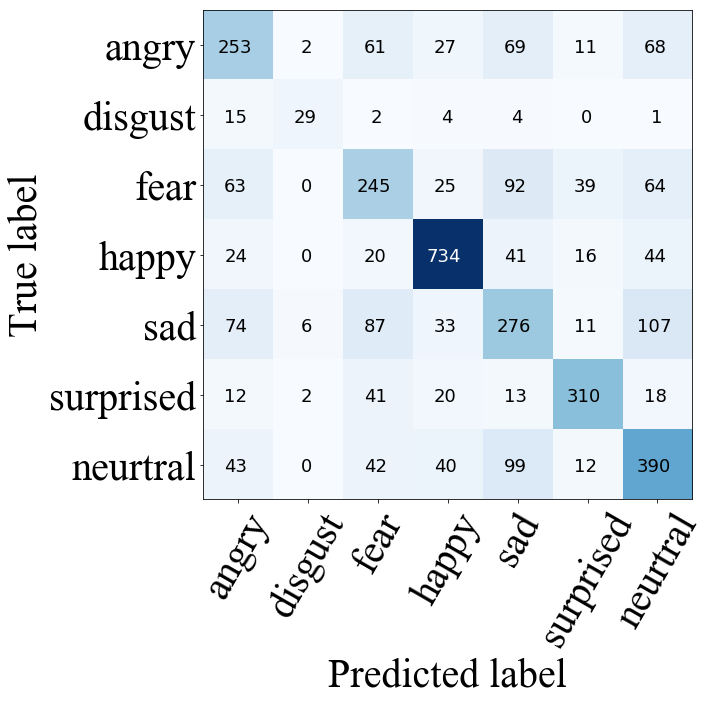}  & \hspace{-0.2in} \includegraphics[width=0.23\textwidth]{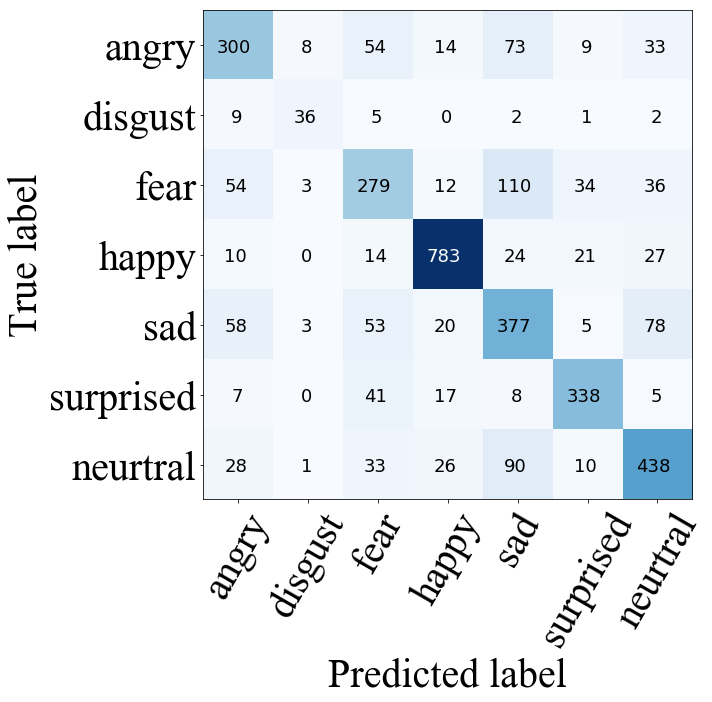}\tabularnewline
\end{tabular}\caption{\label{fig:fer2013_sl_cm}The left image shows the confusion matrix
generated by Comp-GAN-Cls on FER 2013 without pre-training, and the right
one is the Comp-GAN-Cls pre-trained on $\mathrm{F}^{2}$ED. Comp-GAN-Cls refers to the classification backbone of our Comp-GAN.}
\end{figure}

\begin{figure}
\centering{}%
\begin{tabular}{cc}
\hspace{-0.3in}\includegraphics[width=0.23\textwidth]{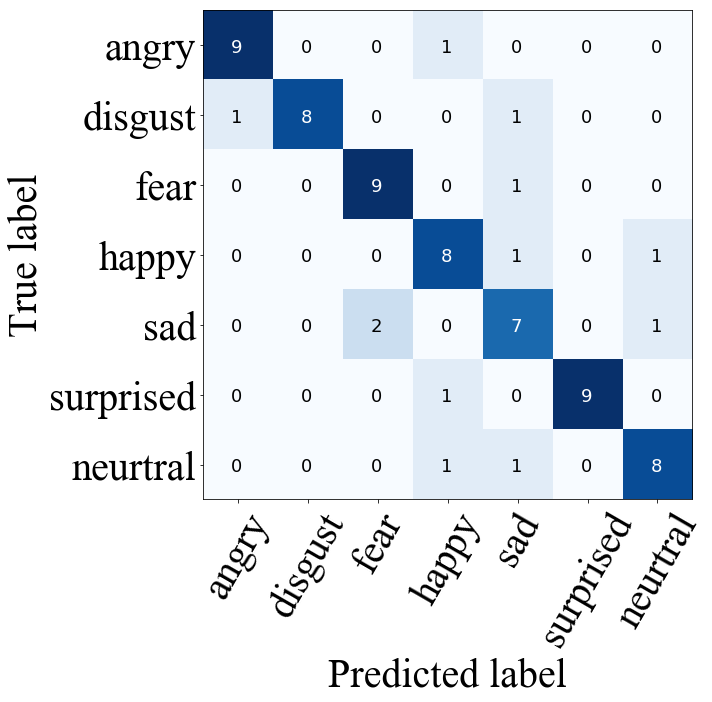}  & \hspace{-0.2in} \includegraphics[width=0.23\textwidth]{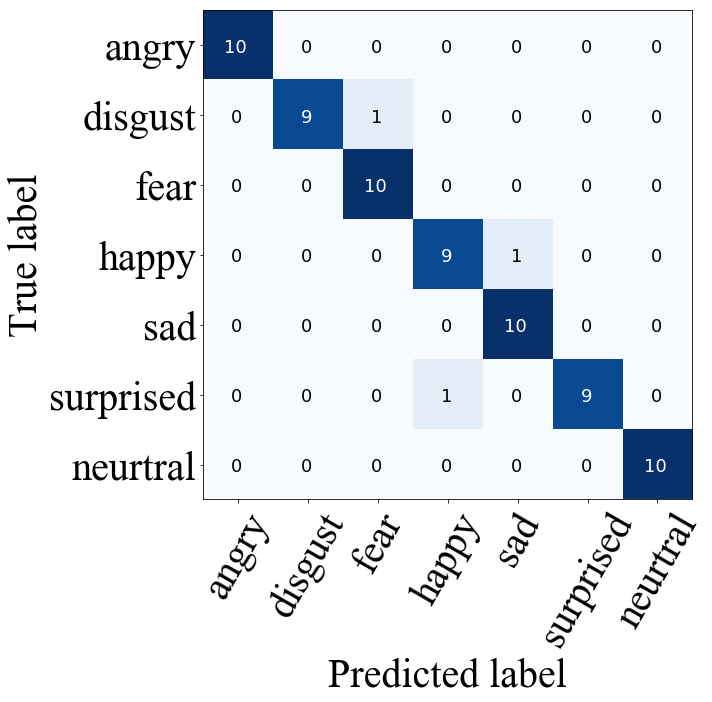}\tabularnewline
\end{tabular}\caption{\label{fig:jaffe_sl_cm}The left image shows the confusion matrix
on FER 2013 for Comp-GAN-Cls without pre-training, and the right one is
the Comp-GAN-Cls pre-trained on $\mathrm{F}^{2}$ED. Comp-GAN-Cls refers to the classification backbone of our Comp-GAN.}
\end{figure}

\subsubsection{Results on supervised learning}

\noindent \textbf{Our dataset $\mathrm{F}^{2}$ED can boost facial
expression recognition accuracy when it is used to pre-train the network 
for better initialization.} To show the efficacy of $\mathrm{F}^{2}$ED,
the classification backbone of Comp-GAN named as Comp-GAN-Cls based on
LightCNN-29v2~\cite{Xiang2015A}
is pre-trained on our dataset which achieves 69.13\% mean accuracy
under the supervised setting, and then fine-tuned on the training set
of FER2013 and JAFFE. Tab.~\ref{tab:Acc_FER2013_SL} and Tab. \ref{tab:Acc_JAFFE_SL}
show that the Comp-GAN-Cls pre-trained on $\mathrm{F}^{2}$ED can improve
the expression recognition performance by 14.5\% and 13.3\% on FER2013 
and JAFFE, respectively, compared
to the one without pre-training.
The confusion matrix in Fig.~\ref{fig:fer2013_sl_cm} shows that
pre-training increases the scores on all expression types of FER2013,
and the confusion matrix in Fig.~\ref{fig:jaffe_sl_cm} shows that
the pre-trained Comp-GAN-Cls only makes 3 wrong predictions and surpasses
the one without pre-training on all expression types of JAFFE. These
demonstrate that the $\mathrm{F}^{2}$ED dataset with large expression and posture
variations can pre-train a deep network with good
initialization parameters. Note that our classification network is
not specially designed for FER task, since it is built upon the 
LightCNN, one typical face recognition architecture.

\noindent \textbf{Our model can achieve the best performance among
competitors.} Compared to the previous methods, the results show that
our model can achieve the accuracy of 76.8\%, which is superior to
the others on FER2013, as compared in Tab.~\ref{tab:Acc_FER2013_SL}.
As listed in Tab.~\ref{tab:Acc_JAFFE_SL}, our model also achieves
the accuracy of 96.2\%, outperforming all the other competitors. Remarkably,
our model surpasses the Attention CNN by 3.4\% in the same data split
setting. The accuracy of CNN+SVM is slightly lower than our model
by 0.9\%, even though their model is trained and tested on the entire
dataset. This shows the efficacy of our dataset in pre-training the
facial expression network.

\noindent 
\begin{table}
\centering{}%
\begin{tabular}{c|c}
\hline 
Model  & Acc.\tabularnewline
\hline 
Bag of Words~ \cite{ionescu2013local}  & 67.4\%\tabularnewline
VGG+SVM~ \cite{georgescu2018local}  & 66.3\%\tabularnewline
Go-Deep\emph{~} \cite{mollahosseini2016going}  & 66.4\%\tabularnewline
DNNRL~ \cite{guo2016deep}  & 70.6\%\tabularnewline
Attention CNN~ \cite{minaee2019deep}  & 70.0\%\tabularnewline
BOVW + local SVM \cite{georgescu2019local}  & 74.9\%\tabularnewline
\hline 
Comp-GAN-Cls w.o Pre-trained  & 62.3\%\tabularnewline
\textbf{Comp-GAN-Cls}  & \textbf{76.8\%}\tabularnewline
\hline 
\end{tabular}
 \caption{\label{tab:Acc_FER2013_SL}Accuracy on FER2013 test set in supervised
setting. Comp-GAN-Cls refers to the classification backbone of our Comp-GAN.}
\end{table}

\begin{table}
\centering{}%
\begin{tabular}{c|c}
\hline 
Model  & Acc.\tabularnewline
\hline 
Fisherface \cite{abidin2012neural}  & 89.2\%\tabularnewline
Salient Facial Patch \cite{happy2015automatic}  & 92.6\%\tabularnewline
CNN+SVM \cite{shima2018image}  & 95.3\%\tabularnewline
Attention CNN \cite{minaee2019deep}  & 92.8\%\tabularnewline
\hline 
Comp-GAN-Cls w.o Pre-trained  & 82.9\%\tabularnewline
\textbf{Comp-GAN-Cls }  & \textbf{96.2\%}\tabularnewline
\hline 
\end{tabular}\caption{\label{tab:Acc_JAFFE_SL}Accuracy on JAFFE test set in supervised
setting.  Comp-GAN-Cls refers to the classification backbone of our Comp-GAN.}
\end{table}

\subsection{Results of Few-shot Expression Recognition}

\subsubsection{Dataset and Setting\label{subsec:Dataset-and-Setting}}

\begin{figure}[t]
\centering{}%
\begin{tabular}{c}
\hspace{-0.25in}\includegraphics[width=1.05\linewidth]{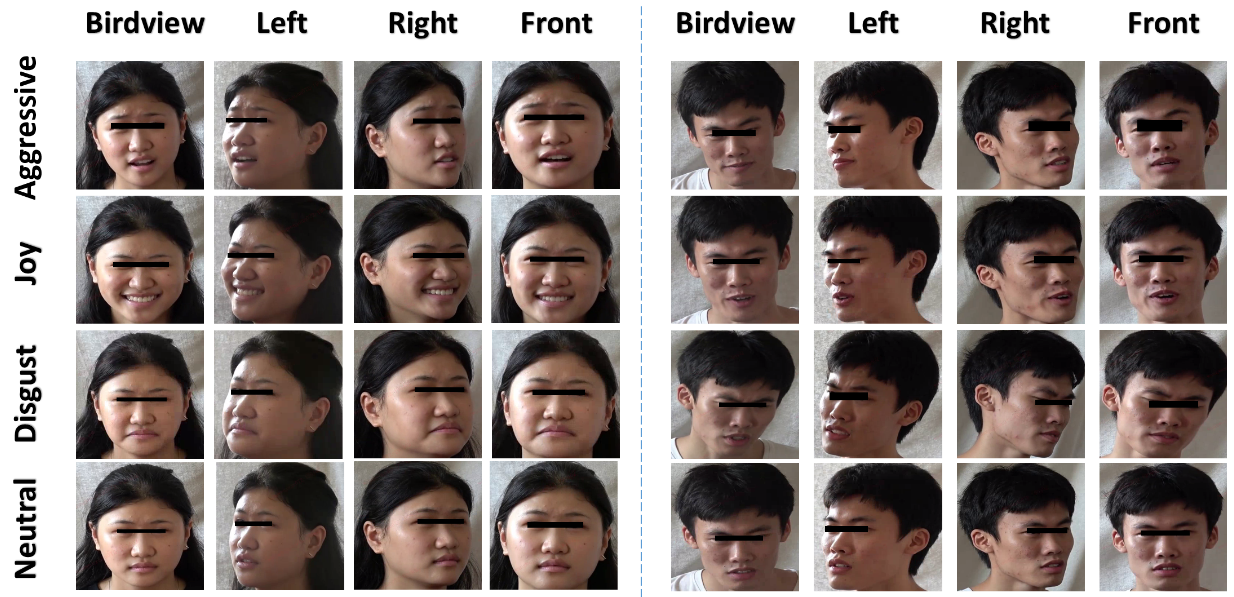}\tabularnewline
\end{tabular}\caption{Faces of different poses/expressions in $\mathrm{F}^{2}$ED.\label{fig:dataset}}
\end{figure}

\noindent \textbf{Fine-Grained Facial Expression Database ($\mathrm{F}^{2}$ED).}
We evaluate our Comp-GAN on $\mathrm{F}^{2}$ED as shown in Fig.~\ref{fig:dataset},
which has 219,719 images with 119 identities and 54 kinds of fine-grained
facial emotions, \textit{e.g.}, acceptance, angry, bravery, calm,
disgust, envy, fear, neutral and so on. Each person is captured by
four different views of cameras, \textit{e.g.}, half left, front, half right,
and bird-view. Each participant expresses his/her emotions under
the guidance of a psychologist, and the images are taken when the
expression is observed. As shown in Fig. \ref{fig:dist_pose} and
Fig. \ref{fig:dist_expression}, $\mathrm{F}^{2}$ED has a relatively
balanced distribution across various expressions as well as postures,
which is beneficial for us to train a generative model that can change
subtle facial expressions and postures synchronously. To ensure the
proportional distribution balance of the test data, we only take 20\%
images from each category into the testing stage on the few-shot
setting, and randomly select 1, 3, or 5 images of $C_{novel}$ and
the rest data of $C_{base}$ images as the training data. For example,
in ER-FID setting, we randomly select 21 identities with 1, 3, or
5 images per person in $D_{novel}$ and 151,189 images of $C_{base}$
in $D_{base}$ for training, and 6,941 images from the 21 identities
$C_{novel}$ and 37,467 images of the remaining persons from $C_{base}$
into the test set. In ER-ZID setting, we randomly choose 20\% of the images
from 21 persons in $C_{novel}$ , \textit{e.g.}, 6,900 images and
the remaining 98 identities $C_{base}$ of 37,861 images into test
set, and 151,444 images of the 98 persons in $D_{base}$ for training.

\noindent \textbf{Implementation details.} We use Pytorch for implementation.
A reasonable architecture for our generators G (Pose) and G (Exp)
is a classic encoder-decoder network~\cite{ronneberger2015u}, which
progressively down-samples the input image into compact hidden space
and then progressively up-samples the hidden information to reconstruct
the image of the same resolution as inputs. Our discriminator follows
the design in AttGAN~\cite{he2019attgan}. Furthermore, our basic classifier is based on LightCNN-29v2~\cite{Xiang2015A}.
For all the experiments, we use the stochastic gradient descent algorithm
to train and dropout is used for fully connected layers with the ratio
0.5. The input images are resized to 144x144 firstly, and then randomly
cropped to 128x128. We pre-train our backbone LightCNN-29v2 on
CelebA dataset \cite{liu2015deep}, set the initial learning rate
of 0.01 and train for 30 epochs. We train the Comp-GAN combined with
LightCNN-29v2 as an end-to-end framework, and set the initial learning
rate as 0.01 in LightCNN-29v2 and 0.0002 for the Comp-GAN. The
learning rate is gradually decreased to zero from the 30th epoch,
and stopped after the 50th epoch. We set the mini-batch size as 64,
$\gamma_{1}=1$, $\gamma_{2}=0.1$, $\lambda_{exp}=20$, $\lambda_{pose}=20$,
$\lambda_{id}=10$, $\lambda_{con}=30$, $\lambda_{recon}=15$, $\lambda_{loop}=10$,
$\lambda_{adv}=1$ and $\mu_{exp}=30$, $\mu_{pose}=10$, $\mu_{id}=15$,
$\mu_{con}=40$, $\mu_{recon}=15$, $\mu_{loop}=10,$$\mu_{adv}=1$.
Our model is trained by one NVIDIA GeForce GTX 1080 Ti GPU and takes
about 11 GB GPU memory.

\noindent \textbf{Evaluation metrics.} The face expression recognition
task can be taken as the problem of classification tasks. To evaluate
the performance, we select five evaluation metrics.

\noindent (1) As used by \cite{pan2010survey,lin2019improving}, the
standard recognition accuracy of each attribute as well as the label-based
metric mean accuracy that overall attributes are computed to evaluate
our model performance, short in \emph{mA.} and \emph{acc.} respectively.

\noindent (2) Instance-based evaluation can capture better consistency
of prediction on a given image \cite{zhang2014review}, to appropriately
evaluate the quality of different methods, following the evaluation
metrics used in pedestrian attribute recognition problem \cite{li2016richly},
we add three more evaluation metrics, \emph{i.e}. precision (\emph{prec}.),
recall (\emph{rec}.) and F1-score (\emph{F1}.).

\noindent (3) Formally, the \emph{acc}., \emph{mA}., \emph{prec}.,
\emph{rec.} and \emph{F1.} can be defined as,

\begin{equation}
mA=\frac{1}{2M}\sum_{i=1}^{M}(TP_{i}/P_{i}+TN_{i}/N_{i})
\end{equation}

\begin{equation}
acc=\frac{1}{M}\sum_{i=1}^{M}(\mid Y_{i}\cap f(x_{i})\mid/\mid Y_{i}\cup f(x_{i})\mid)
\end{equation}

\begin{equation}
prec=\frac{1}{M}\sum_{i=1}^{M}(\mid Y_{i}\cap f(x_{i})\mid/\mid f(x_{i})\mid)
\end{equation}

\begin{equation}
rec=\frac{1}{M}\sum_{i=1}^{M}(\mid Y_{i}\cap f(x_{i})\mid/\mid Y_{i}\mid)
\end{equation}

\begin{equation}
F1=(2\times prec\times rec)/(prec+rec)
\end{equation}

\noindent where $M$ is the total number of attributes; $P_{i}$ and
$TP_{i}$ are the numbers of positive examples and correctly predicted
positive examples; $N_{i}$ and $TN_{i}$ are the numbers of negative
examples and correctly predicted negative examples. $Y_{i}$ is the
ground truth positive labels of the $i-th$ example, $f(x_{i})$ returns
the predicted positive labels for $i-th$ example, and $\mid\cdot\mid$
means the set cardinality.

\subsubsection{Results of Comp-GAN Vs. Competitors\label{subsec:Ablation-Study}}

\noindent We compare our model against the existing generative model,
such as Cycle-GAN~\cite{zhu2017unpaired}, Pix2Pix~\cite{isola2017image},
VAE/GAN\cite{larsen2015autoencoding} and AttGAN~\cite{he2019attgan}.
In particular, we highlight the following observations,

\begin{figure}[t]
\centering{}%
\begin{tabular}{c}
\hspace{-0.25in}\includegraphics[width=1.05\linewidth]{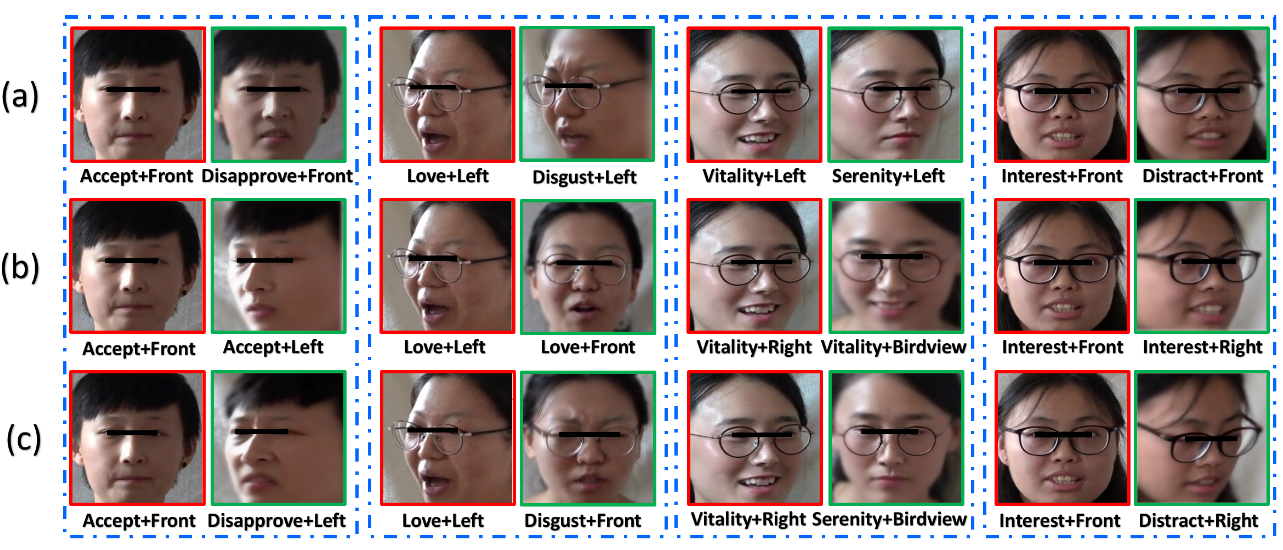}\tabularnewline
\end{tabular}\caption{Generated samples from our Comp-GAN. The images with red borders represent
the input references, and the images with green borders are the synthesized
faces. (a) shows the images generated by generator G (Exp) which changes
the expression while keeping the pose and identity information; images
in (b) are generated from generator G (Pose) to edit posture; (c)
illustrates the images with desired expression and specified pose
through our Comp-GAN model. The images belong to the same identity
circled by the blue dash line. \label{fig:generated_images}}
\end{figure}

\vspace{0.1in}
\noindent \textbf{Quality Comparison. }

\noindent \textbf{\textit{(1) Comp-GAN can well edit the facial images.}}
As in Fig.~\ref{fig:generated_images}, we show several realistic images with
the specified pose, desired expression generated by Comp-GAN
on $\mathrm{F}^{2}$ED. We notice that the images have dramatically
changed postures and expressions, and they can still maintain the
expression and identity while changing the posture, and vice versa.

\begin{figure}[h]
\centering{}%
\begin{tabular}{c}
\hspace{-0.05in} \includegraphics[width=0.95\linewidth]{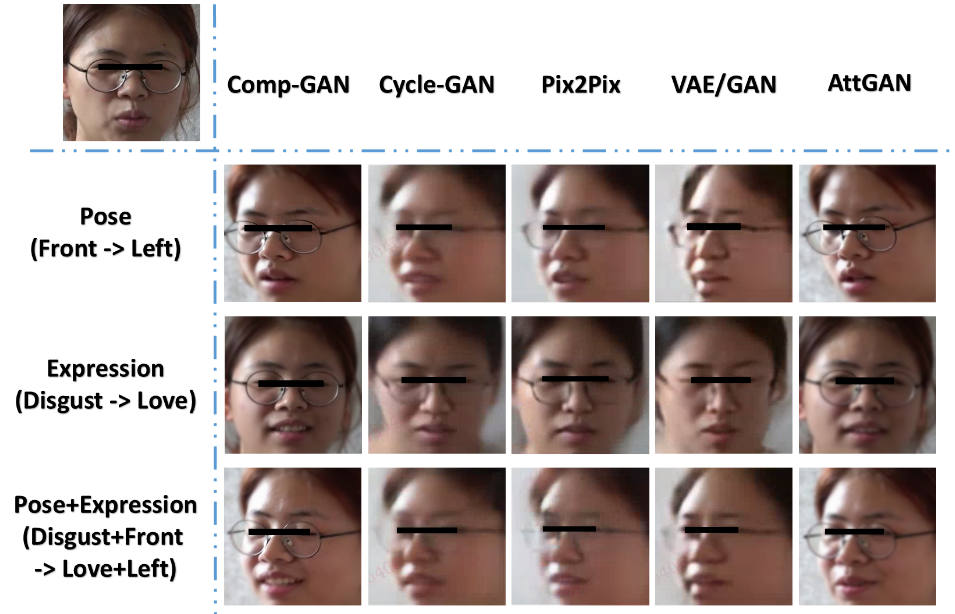}\tabularnewline
\end{tabular}\caption{Comparison results with other generative methods. The input reference
image is at the upper left corner. The first line shows the generated
images from generator G (Pose), the second line generated by generator
G (Exp), and the last line illustrates the images with desired expression
and specific pose. \label{fig:GAN_Comparsion}}
\end{figure}

\noindent \textbf{\textit{(2) Comp-GAN can generate more realistic
faces compared to other methods.}} As shown in Fig.~\ref{fig:GAN_Comparsion},
our Comp-GAN model can preserve better face identity. In contrast,
the generated images from Cycle-GAN, Pix2Pix, and VAE/GAN do not well
edit facial expression and retain the original identity. Their methods
also lose some other facial attributes, such as the hairstyle and
glasses, and the quality of produced images are worse. As we can see,
both Comp-GAN and AttGAN accurately generate the desired expression
while keeping the expression-excluding information, but the AttGAN
results contain some artifacts and are much blurrier than ours, while
the images from Comp-GAN seem more natural and realistic.

\begin{table}[t]
\centering%
\begin{tabular}{c}
\hspace{-0.1in}%
\begin{tabular}{lcccccc}
\toprule 
\hspace{-0.05in}Method  & \hspace{-0.05in}ER-SS  & \hspace{-0.05in}ER-FID  & \hspace{-0.05in}ER-ZID  & \hspace{-0.05in}ER-FP  & \hspace{-0.05in}EP-ZP  & \hspace{-0.05in}ER-FE\tabularnewline
\midrule 
\hspace{-0.05in}L.  & \hspace{-0.05in}70.63  & \hspace{-0.05in}33.29  & \hspace{-0.05in}10.44  & \hspace{-0.05in}49.32  & \hspace{-0.05in}42.17  & \hspace{-0.05in}38.94\tabularnewline
\hspace{-0.05in}L. + Cycle-GAN  & \hspace{-0.05in}65.38  & \hspace{-0.05in}29.34  & \hspace{-0.05in}4.02  & \hspace{-0.05in}50.19  & \hspace{-0.05in}42.29  & \hspace{-0.05in}30.68\tabularnewline
\hspace{-0.05in}L. + Pix2Pix  & \hspace{-0.05in}71.31  & \hspace{-0.05in}31.68  & \hspace{-0.05in}7.93  & \hspace{-0.05in}43.12  & \hspace{-0.05in}34.75  & \hspace{-0.05in}37.25\tabularnewline
\hspace{-0.05in}L. + VAE/GAN  & \hspace{-0.05in}70.88  & \hspace{-0.05in}25.10  & \hspace{-0.05in}5.17  & \hspace{-0.05in}37.34  & \hspace{-0.05in}30.89  & \hspace{-0.05in}33.96\tabularnewline
\hspace{-0.05in}L. + AttGAN  & \hspace{-0.05in}71.92  & \hspace{-0.05in}34.64  & \hspace{-0.05in}10.98  & \hspace{-0.05in}50.14  & \hspace{-0.05in}44.81  & \hspace{-0.05in}40.14\tabularnewline
\midrule 
\hspace{-0.05in}\textbf{Comp-GAN}  & \hspace{-0.05in}\textbf{74.92}  & \hspace{-0.05in}\textbf{36.92}  & \hspace{-0.05in}\textbf{14.25}  & \hspace{-0.05in}\textbf{56.43}  & \hspace{-0.05in}\textbf{51.29}  & \hspace{-0.05in}\textbf{44.19}\tabularnewline
\bottomrule
\end{tabular}\tabularnewline
\end{tabular}\caption{Comparison results with other generative frameworks on F$^{2}$ED.
L.: indicates the facial expression recognition backbone LightCNN. Only 1 training
image of each class in $C_{novel}$ under ER-FID, ER-FP, and ER-FE
settings. \label{tab:other_GAN_accuracy}}
\end{table}

\begin{table*}[t]
\centering{}%
\begin{tabular}{lcccccc}
\toprule 
Method  & ER-SS  & ER-FID  & ER-ZID  & ER-FP  & EP-ZP  & ER-FE\tabularnewline
\midrule 
L.  & 70.63  & 33.29  & 10.44  & 49.32  & 42.17  & 38.94\tabularnewline
L.+ G (Pose)  & 73.11  & 35.56  & 12.31  & 54.49  & 47.92  & 40.93\tabularnewline
L.+ G (Pose) + Front  & 73.49  & 35.89  & 13.04  & 54.92  & 48.87  & 41.76\tabularnewline
L.+ G (Pose) + Combine  & 73.84  & 35.92  & 13.48  & 55.43  & 49.02  & 41.91\tabularnewline
L.+ G (Exp)  & 72.36  & 34.95  & 12.87  & 53.21  & 43.98  & 42.97\tabularnewline
L.+ G (Pose)+ G (Exp)  & 74.03  & 36.78  & 13.98  & 56.04  & 50.57  & 43.84\tabularnewline
\midrule 
\textbf{Comp-GAN}  & \textbf{74.92}  & \textbf{36.92}  & \textbf{14.25}  & \textbf{56.43}  & \textbf{51.29}  & \textbf{44.19}\tabularnewline
\bottomrule
\end{tabular}\hspace{-0.3in} \caption{Results on F$^{2}$ED. Chance-level =2\%. Note that (1) L.: indicates
the facial expression backbone recognition LightCNN-29v2; (2) Only 1 training
image of each class in $C_{novel}$ on ER-FID, ER-FP, and ER-FE. \label{tab:self-evaluation}}
\end{table*}

\vspace{0.1in}
\noindent \textbf{Quantity Comparison.}

\noindent \textbf{\textit{(1) Comp-GAN achieves the best performance
among the competitors.}} We use the same number of generated images
from AttGAN, Cycle-GAN, Pix2Pix, VAE / GAN, and AttGAN to fine-tune
the same feature extractor respectively. As in Tab.~\ref{tab:other_GAN_accuracy}
our model gains the highest accuracy under all proposed settings on
F$^{2}$ED, and some generated images of other models even damage
the recognition accuracy, such as Cycle-GAN and VAE/GAN under the
few-shot setting. This well proves that Comp-GAN generated images
can solve the problem of the lack of data, limitation in diversity
and huge changes of posture for the facial expression recognition
task.

\noindent \textbf{\textit{(2) The generated faces from Comp-GAN preserve
the most accurate original image information.}}
To further demonstrate the accuracy of our generated images
in expression, posture, and identity, we employ a trained classifier
of expression, pose and identity as the measurement tool. The classifier
is learned on $\mathrm{F}^{2}$ED training set, and it can thus achieve
85.32\% accuracy per attribute, 97.69\% per pose and 98.63\% per identity
on the training images of $\mathrm{F}^{2}$ED. If the category of
a generated image is predicted the same as the desired one by the
classifier, it should be considered as the correct generation on this dataset.

\begin{figure}[h]
\centering{}%
\begin{tabular}{c}
\hspace{-0.1in}\includegraphics[width=1\columnwidth]{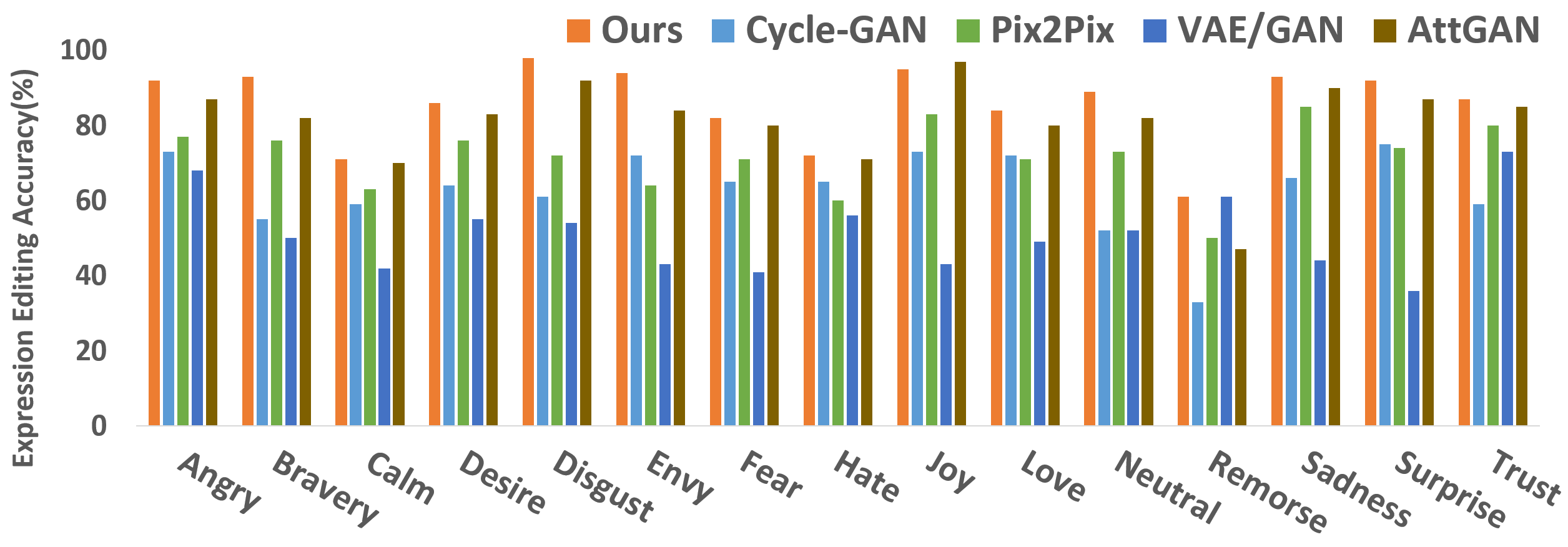}\tabularnewline
\end{tabular}\caption{Comparisons among other generative models and Comp-GAN in terms of
facial expression preserving accuracy. (the higher, the better).\label{fig:expression_editing_accuracy}}
\end{figure}

As in Fig.~\ref{fig:expression_editing_accuracy}, we show
the expression editing accuracy of the randomly selected 15 kinds
of desired expression generated images, and compare them with
other generative results. We can notice that Comp-GAN and AttGAN achieve
much better accuracies than Cycle-GAN, Pix2Pix and VAE/GAN among all
the expressions. As for the comparisons between the Comp-GAN and AttGAN,
AttGAN can achieve better performance with a slight margin on `Joy'
expression, however, our model can get superior accuracy on the rest
14 expressions, and the generated images of Comp-GAN are much more
natural and realistic as shown in Fig.~\ref{fig:GAN_Comparsion}.

\begin{figure}[h]
\centering{}%
\begin{tabular}{c}
\hspace{-0.1in}\includegraphics[width=1\columnwidth]{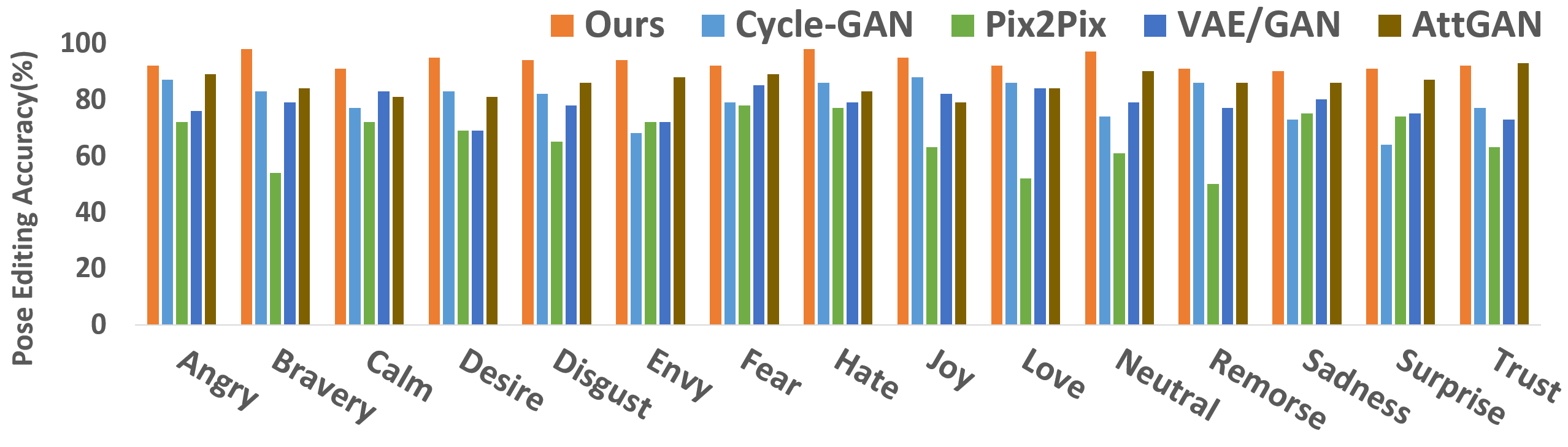}\tabularnewline
\end{tabular}\caption{Comparisons among other generative models and Comp-GAN in terms of
pose preserving accuracy. (the higher, the better).\label{fig:pose_editing_accuracy}}
\end{figure}

And we also compare with other generative results and show
the pose accuracy using 4 kinds of specified pose generated images in
Fig.~\ref{fig:pose_editing_accuracy}. In contrast to the complex
and varied expression modifications, the generated images from all
methods are more accurate in poses, and Comp-GAN achieves the best
performance on the posture accuracy. We can still see that our method
can more realistically retain the original identity and expression
information in the case of posture change from Fig.~\ref{fig:GAN_Comparsion}.

\begin{figure}[h]
\centering{}%
\begin{tabular}{c}
\hspace{-0.1in}\includegraphics[width=1\columnwidth]{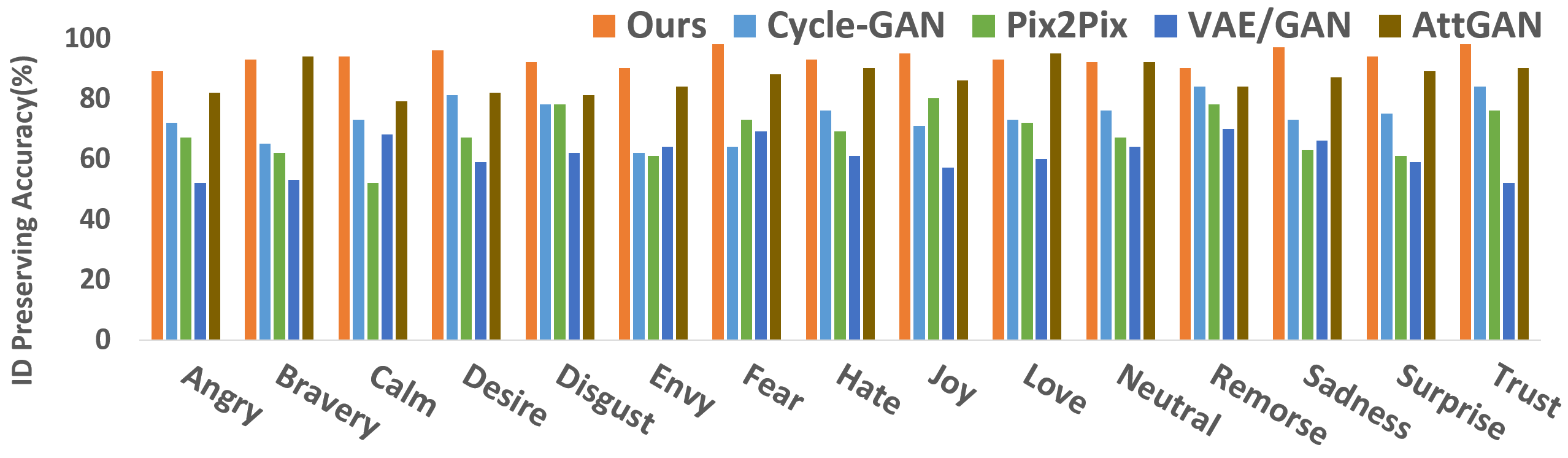}\tabularnewline
\end{tabular}\caption{Comparisons among other generative models and Comp-GAN in terms of
identity preserving accuracy. (the higher, the better).\label{fig:identity_editing_accuracy}}
\end{figure}

Furthermore, we analyze the identity preserving accuracy
of all generated images shown in Fig.~\ref{fig:identity_editing_accuracy}.
We can find that, in comparison, AttGAN and Comp-GAN are more accurate
in the preservation of identity information, and our method gets higher
accuracy for most attributes, which can be attributed to the well-designed
generation network with reconstruction structure and loss function.

\subsection{Evaluating Each Component of Comp-GAN}

\noindent We conduct extensive qualitative and quantitative experiments
to evaluate the effectiveness of each component in Comp-GAN.

\vspace{0.1in}
\noindent \textbf{Quantity results.} To make quantitative self-evaluations,
we conduct extensive experiments on several variants: (1) LightCNN:
we use the $\mathrm{F}^{2}$ED training data to fine-tune the LightCNN-29v2
model which is pre-trained on CelebA. (2) LightCNN + G (Pose): we
add the pose changed images synthesized from generator G (Pose)
to $\mathrm{F}^{2}$ED training set and fine-tune the LightCNN-29v2
model. (3) LightCNN + G (Pose) + Front: we not only use the specified-pose
generated images, but also transfer all the training and testing data
by generator G (Pose) into the front pose to extract features. (4)
LightCNN + G (Pose) + Combine: we classify all the training and testing
data into four poses, \textit{e.g.}, front, left, right and bird-view,
and we concatenate the features extracted from those four kinds of
generated images with the input one extracted by LightCNN-29v2 as
the final feature. (5) LightCNN + G (Exp): we add the desired expression
synthesized images from generator G (Exp) to $\mathrm{F}^{2}$ED training
set and fine-tune the LightCNN-29v2 model. (6) LightCNN + G (pose)
+ G (Exp): we add the specified pose synthesized images and the desired
expression generated images to $\mathrm{F}^{2}$ED training set and
fine-tune the LightCNN-29v2 model. (7) Ours (LightCNN + Comp-GAN):
we add the specified pose and desired expression generated images
into the training set, transfer all the training and testing data into four
poses, and concatenate the five features as the final output.

\noindent As in Tab.~\ref{tab:self-evaluation}, we compare the recognition
accuracy of the variants of our model under six settings, as follows,

\noindent \textbf{\textit{(1) The efficacy of our generators.}} Our
`LightCNN + G (Pose)' and `LightCNN + G (Exp)' greatly outperforms
`LightCNN' in Tab.~\ref{tab:self-evaluation} under all six settings.
Especially in ER-FP and EP-ZP setting, `LightCNN + G (Pose)' can achieve
5.17\% and 5.75\% improvement respectively; in ER-FE setting, `LightCNN
+ G (Exp)' shows 4.03\% improvement; as for ER-FID and ER-ZID setting,
our `LightCNN + G (Pose)' and `LightCNN + G (Exp)' can boost the recognition
accuracy by 2\% on average. Our generative method achieves superior
performance, meaning that our generated faces can increase data diversity
and preserve useful information.

\noindent \textbf{\textit{(2) The complementary property of features between our generators
.}} To be noticed, `LightCNN + G (Pose)' can obtain
better performance on posture-limited data than `LightCNN + G (Exp)',
and `LightCNN + G (Exp)' has superior accuracy over `LightCNN + G
(Pose)' in the expression-limited setting, which indicates that these
two generators have different emphasis on the expression recognition
task. As can be seen, `LightCNN + G (Pose) + G (Exp)' model can beat
`LightCNN + G (Pose)' and `LightCNN + G (Exp)' with the visible margin
under all six settings, and this strongly proves the complementarity
features between our specified-pose and desired-expression generated
data can help the network learn better facial expression representation.

\noindent \textbf{\textit{(3) The importance of facial pose normalization.}}
Furthermore, our `LightCNN + G (Pose) + Front' and `LightCNN + G (Pose)
+ Combine' variants achieve better performance than `LightCNN + G
(Pose)', so such multiple poses normalization method is an effective
way to deal with the large pose variation problem in facial expression
recognition task.

\begin{figure}[t]
\centering{}\includegraphics[width=1\linewidth]{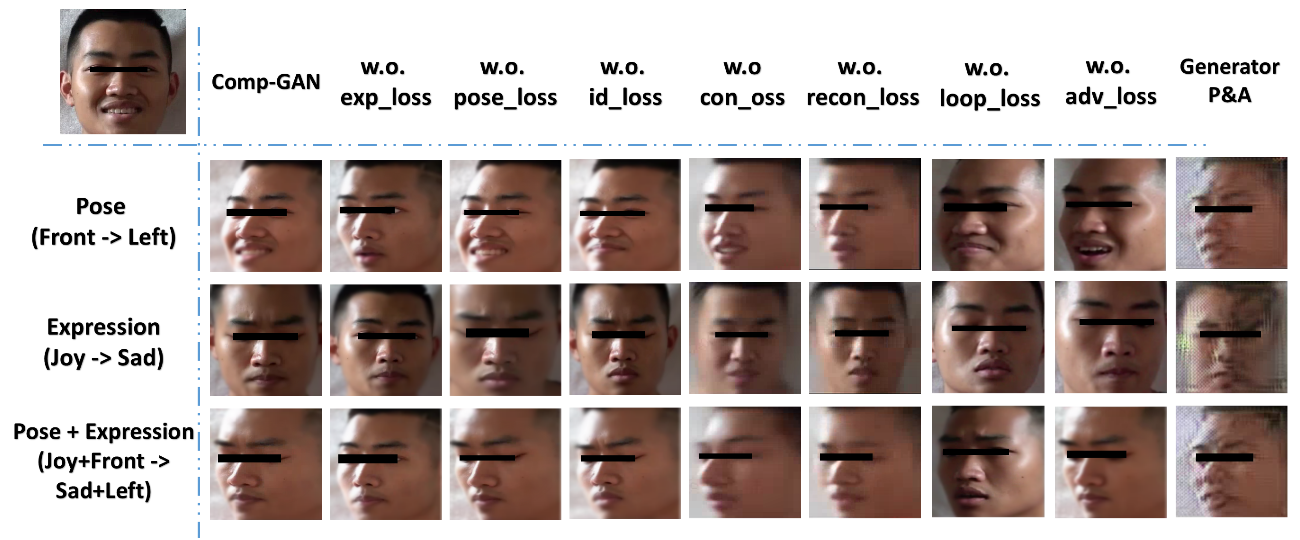}\caption{Ablation study of loss functions in Comp-GAN. The input reference
image is at the upper left corner. The first row shows generated images
from generator G (Pose); the second row generated form generator G
(Exp); and the last line illustrates the images with both expressions
and poses changed. w.o.: indicates that we remove this loss function
during the training.\label{fig:GAN_Loss}}
\end{figure}

\begin{table*}
\centering{}%
\begin{tabular}{lccccccccc}
\toprule 
Method  & \multicolumn{3}{c}{ER-FID} & \multicolumn{3}{c}{ER-FP} & \multicolumn{3}{c}{ER-FE}\tabularnewline
\midrule 
 & 1 Shot  & 3 Shot  & 5 Shot  & 1 Shot  & 3 Shot  & 5 Shot  & 1 Shot  & 3 Shot  & 5 Shot\tabularnewline
\midrule 
L.  & 33.29  & 33.45  & 34.36  & 49.32  & 49.41  & 50.21  & 38.94  & 39.02  & 39.87\tabularnewline
L. + G (Pose)  & 35.56  & 35.74  & 36.27  & 54.49  & 54.44  & 55.72  & 40.93  & 40.04  & 41.26\tabularnewline
L. + G (Exp)  & 34.95  & 35.81  & 37.01  & 52.21  & 52.89  & 53.24  & 42.97  & 41.86  & 42.50\tabularnewline
L. + G (Pose) + G(Exp)  & 36.78  & 36.88  & 37.43  & 56.04  & 56.56  & 57.36  & 43.84  & 43.90  & 44.98\tabularnewline
\midrule 
\textbf{Comp-GAN}  & \textbf{36.92}  & \textbf{36.98}  & \textbf{37.69}  & \textbf{56.43}  & \textbf{56.58}  & \textbf{57.59}  & \textbf{44.19}  & \textbf{44.22}  & \textbf{45.67}\tabularnewline
\bottomrule
\end{tabular}\caption{Results of few-shot setting on F$^{2}$ED. L.: indicates the facial
expression recognition backbone LightCNN-29v2.\label{tab:one-shot setting}}
\end{table*}

\vspace{0.1in}
\noindent \textbf{\textit{\emph{Effectiveness of loss function in
Comp-GAN.}}} To generate realistic and information-keeping images,
we apply six loss functions in Comp-GAN: (a) expression-prediction
loss (exp\_loss), (b) posture-prediction loss (pose\_loss), (c) ID-preserving
loss (id\_loss), (d) construction loss (con\_loss), (e) reconstruction
loss (recon\_loss), (f) closed-loop loss (loop\_loss), (g) adversarial loss (adv\_loss). To verify the
effectiveness of these losses, we conduct experiments by removing
each of them respectively during the training and present the generated
results in Fig.~\ref{fig:GAN_Loss}. As can be seen, without the loss
(a), the expression information is severely lost, and the model tends
to generate natural facial expression; without (b) or (c), the quality
of the generated images is degraded and the faces become blurred.
Without loss (d) or (e), the generated image quality becomes poor
and lacks vital identity information, showing that the two loss
functions are essential for our Comp-GAN. By adding the loss (f), 
Comp-GAN can greatly improve the quality of the generated images.
Removing the loss function (g), the quality of synthetic faces all degrades to different extent.
We also try to add the pose information as a part of expression, and
train only one generator (Generator P\&A) to edit posture and expression
simultaneously, as shown in Fig.~\ref{fig:GAN_Loss}, the generator
is not conducted at all. This is the underlying reason we designed
the two generators that edit pose and expression separately.

\subsection{Ablation Study }

\begin{figure}
\begin{centering}
\begin{tabular}{c}
\includegraphics[width=0.45\textwidth]{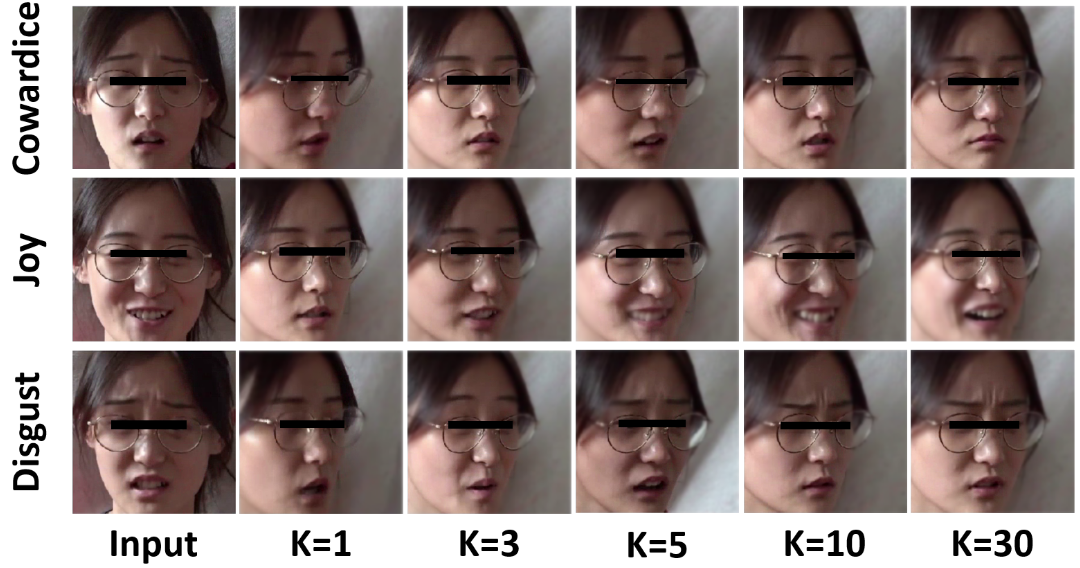}\tabularnewline
\end{tabular}
\par\end{centering}
\caption{\label{fig:few-shot generation}Generated images with different $k$
training examples for each $C_{novel}$ under few-shot setting on
$\mathrm{F}^{2}$ED.}
\end{figure}

\noindent \textbf{Quality of synthesized images in terms of the amount
of training data under the few-shot setting.} As shown in Fig. \ref{fig:few-shot generation},
we show the synthesized images by our Comp-GAN when the number of
training images $k$ per $C_{novel}$ expression category varies on
F$^{2}$ED. We can notice that the quality of the generated images
is gradually improved as $k$ increases. When the training number
$k$ is 1, the generated faces are closer to natural expression,
this indicates that due to the lack of training data, the generator
do not learn the relevant feature of specific facial expressions well.
However, when the training number $k$ is more than 5, our Comp-GAN
can extract specific expression representation and generate more realistic
faces, which maintain the same identity and expression information
as the input while change the pose.

\noindent \textbf{Our model can solve the problem of insufficient
training data such as few-shot learning setting.} We show the accuracy
results of the few-shot setting on F$^{2}$ED in Tab.~\ref{tab:one-shot setting},
and Comp-GAN can significantly improve the expression recognition
accuracy compared to the baseline `LightCNN'. It is obvious that,
with the number of training images $k$ per $C_{novel}$ category
increasing, the recognition accuracy also gradually improved. Compared
with `LightCNN + G (Exp)' and `LightCNN + G (Pose)', the `LightCNN
+ G (Exp)' is more effective for the lack of expression data, and
`LightCNN + G (Pose)' can greatly improve the recognition accuracy
for the data with fewer postures changed, and `LightCNN + G (Pose)
+ G (Exp)' achieves better performance than `LightCNN + G (Exp)' or
`LightCNN + G (Pose)' model. This further demonstrates the different
focuses and complementary properties of the two kinds of generated data.
The promising results show the strong generalization capability of our
method and efficacy in real applications.

\begin{table}
\centering{}%
\begin{tabular}{c}
\hspace{-0.1in}%
\begin{tabular}{lccccc}
\toprule 
Method  & Acc.  & mA.  & Pre.  & Rec.  & F1.\tabularnewline
\midrule 
L.  & 38.94  & 36.83  & 40.35  & 29.32  & 33.96\tabularnewline
L.+ G(Pose)  & 40.93  & 41.24  & 44.78  & 31.45  & 36.95\tabularnewline
L.+ G(Exp)  & 42.97  & 42.98  & 42.13  & 30.17  & 35.16\tabularnewline
L.+ G(pose) + G(Exp)  & 43.84  & 44.46  & 48.52  & 32.80  & 39.14\tabularnewline
L.+ AttGAN  & 40.14  & 38.55  & 43.90  & 26.55  & 33.09\tabularnewline
\midrule 
\textbf{Comp-GAN}  & \textbf{44.19}  & \textbf{45.79}  & \textbf{50.18}  & \textbf{35.02}  & \textbf{41.25}\tabularnewline
\bottomrule
\end{tabular}\tabularnewline
\end{tabular}\caption{Comparison results of ER-FE setting with 1 training
image of each class in $C_{novel}$ on F$^{2}$ED. L.: indicates the
facial expression recognition backbone LightCNN-29v2. \label{tab:metrics}}
\end{table}

\noindent \textbf{Our model can achieve the best performance among
five metrics.} Tab. ~\ref{tab:metrics} shows the results of our
model, its variants and the most comparable existing generative method
AttGAN~\cite{he2019attgan} among five metrics under ER-FE setting
in F$^{2}$ED. As can be seen, our model gets the best results and
beats the AttGAN with a noticeable gap. It is worth noticing that
among five metrics, `LightCNN + G (Exp)' and `LightCNN + G (Pose)'
models have better performance compared with `LightCNN', this further
proves the validity of our generated images. On the other settings,
we have similar observations.

\begin{figure}[t]
\centering{}\includegraphics[width=0.98\linewidth]{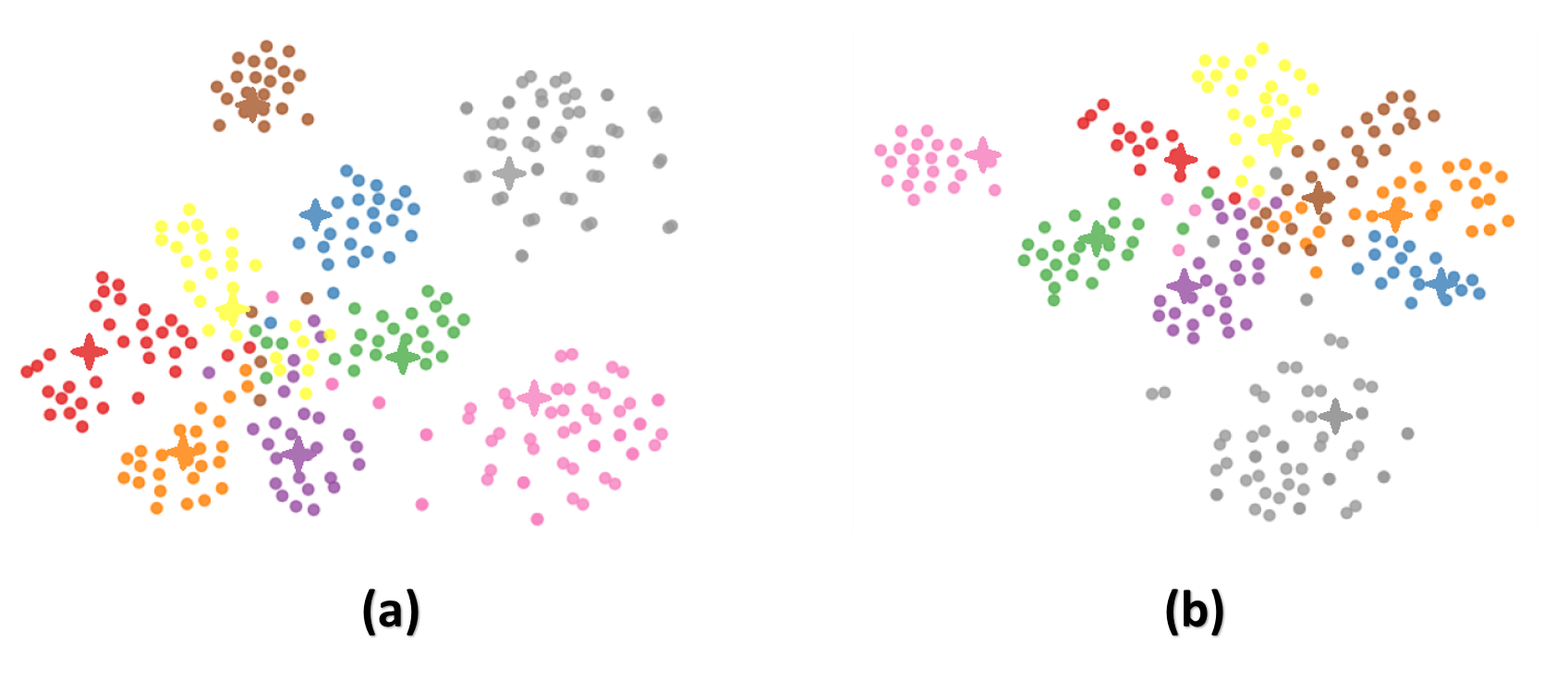}
\hspace{-0.2in} \caption{Visualization of 9 original images (drawn as stars) and the corresponding
generated images (drawn as dots) using t-SNE. One color indicates
one identity. (a) and (b) show the image identity distribution generated
by G (Pose) or G (Exp) respectively. Best view in color. \label{fig:T-sne}}
\end{figure}

\noindent \textbf{\textit{\emph{Quality of synthesized images.}}}
For implications in real-world applications, we expect the generated
faces not only to look realistic but also preserve identity information.
As shown in Fig.~\ref{fig:T-sne}, it visualizes the identity feature
distributions of original and specified pose or desired expression
generated data using randomly sampled 9 images via t-SNE. One color
indicates one identity, and it is noticeable that the generated data
are clustered around the original images with the same identities.
It means our Comp-GAN can effectively preserve the identity information
during the generative process.

\noindent \textbf{The number of synthesized images.} 
We choose to generate 10 synthesized images for each novel input category in the
former experiments. To evaluate the relationship between the number
of generated images and the recognition accuracy under the few-shot
learning task, we also compare the results of generating 0, 1, 3,
5, 10, 20, 50, 100, 200, 500, 1000 synthesized images, while all the
other parameters are kept the same ( the number of training images
$k$ per $C_{novel}$ category is set as $k=5$). Under the ER-FE
setting, we list the corresponding expression recognition mean accuracy
as: 39.24\%, 44.19\%, 44.22\%, 45.67\%, 47.45\%, 49.78\%, 53.98\%,
55.17\%, 56.31\%, 56.92\%, and 57.14\%, respectively. It's clear that
changing this parameter may lead to a slight change in the final performance,
but our results are still significantly better than the baseline.

\section{Conclusion}

In this work, we introduce $\mathrm{F}^{2}$ED, a new facial expression
database containing 54 different fine-grained expression types and
more than 200k examples. This largely complements the lack of diversity
in the existing expression datasets. Furthermore, we propose a novel
end-to-end compositional generative adversarial network (Comp-GAN)
framework to generate natural and realistic face images and we use
the generated images to train a robust expression recognition model.
Comp-GAN can dynamically change the facial expression and pose according
to the input reference images, while preserving the expression-excluding
details. To evaluate the framework, we perform several few-shot learning
tasks on $\mathrm{F}^{2}$ED dataset, and the results show that our
model can relieve the limitation of data. Subsequently, we fine-tune
our model pre-trained on $\mathrm{F}^{2}$ED for the existing FER2013
and JAFFE database, and the results demonstrate the efficacy of our
dataset in pre-training the facial expression recognition network.















\ifCLASSOPTIONcaptionsoff \newpage\fi




 \bibliographystyle{ieeetr}
\bibliography{egbib}

%









\end{document}